\pdfoutput=1

\documentclass[11pt]{article}
\PassOptionsToPackage{table}{xcolor}
\usepackage[final]{acl}

\usepackage{times}
\usepackage{latexsym}
\usepackage{graphicx}  
\usepackage{xspace}
\usepackage{multirow} 
\usepackage{booktabs} 
\usepackage{arydshln} 
\usepackage{makecell}
\usepackage{amsmath}
\usepackage{amssymb}

\usepackage[table]{xcolor} 
\usepackage[most]{tcolorbox}
\definecolor{SkyBlue}{rgb}{0.88, 0.94, 1.0}
\definecolor{Purple}{HTML}{7030a0}
\definecolor{Blue}{HTML}{4472c4}
\definecolor{Green}{HTML}{119f40}
\usepackage{enumitem} 
\definecolor{lightblue}{HTML}{dceafa}    
\newcommand{\myblue}{\cellcolor{lightblue}}

\usepackage[T1]{fontenc}

\usepackage[utf8]{inputenc}

\usepackage{microtype}

\usepackage{inconsolata}

\usepackage{graphicx}
\usepackage{booktabs}
\usepackage{subcaption}
\newcommand{\ours}{GSI\xspace}

\makeatletter
\def\@fnsymbol#1{\ensuremath{\ifcase#1\or *\or  \ddagger\or \dagger\or
   \mathsection\or \mathparagraph\or \|\or **\or \dagger\dagger
   \or \ddagger\ddagger \else\@ctrerr\fi}}
\makeatother

\title{Mitigating Tail Narrowing in LLM Self-Improvement \\ via Socratic-Guided Sampling}

\author{
\textbf{
Yiwen Ding$^{1}$\thanks{\ Equal contribution. $^\ddag$ Work done during internship at Meituan. $^\dag$ Corresponding author.}$^{\ddag}$, {}
Zhiheng Xi$^1$\footnotemark[1],  { }
Wei He$^1$, { }
Zhuoyuan Li$^5$, { }
Yitao Zhai$^2$, { }
}
\\
\textbf{
Xiaowei Shi$^2$, { }
Xunliang Cai$^2$, { }
Tao Gui$^{3\dag}$,  { }
Qi Zhang$^{1,4}$,  { }
Xuanjing Huang$^{1,4\dag}$}
\\
  {$^1$ School of Computer Science, Fudan University} \ \ 
  {$^2$ Meituan} \\
  {$^3$ Institute of Modern Languages and Linguistics, Fudan University} \\
  {$^4$ Key Laboratory of Intelligent Information Processing, Fudan University} \\
  {$^5$ Macau University of Science and Technology}\\
    \texttt{
    \{ywding23,zhxi22\}@m.fudan.edu.cn, \{tgui, xjhuang\}@fudan.edu.cn}
}

\begin{document}
\maketitle
\begin{abstract}

Self-improvement methods enable large language models (LLMs) to generate solutions themselves and iteratively train on filtered, high-quality rationales.
This process proves effective and reduces the reliance on human supervision in LLMs' reasoning, but the performance soon plateaus.
We delve into the process and find that models tend to over-sample on easy queries and under-sample on queries they have yet to master. 
As iterations proceed, this imbalance in sampling is exacerbated, leading to a long-tail distribution where solutions to difficult queries almost diminish.
This phenomenon limits the performance gain of self-improving models.
A straightforward solution is brute-force sampling to balance the distribution, which significantly raises computational costs.
In this paper, we introduce \textbf{G}uided \textbf{S}elf-\textbf{I}mprovement (\textbf{\ours}), a strategy aimed at improving the efficiency of sampling challenging heavy-tailed data. 
It leverages Socratic-style guidance signals to help LLM reasoning with complex queries, reducing the exploration effort and minimizing computational overhead.
Experiments on four models across diverse mathematical tasks show that \ours strikes a balance between performance and efficiency, while also being effective on held-out tasks\footnote{\ Codes are publicly available at \href{https://github.com/Yiwen-Ding/Guided-Self-Improvement}{https://github.com/Yiwen-Ding/Guided-Self-Improvement}.}.

\end{abstract}

\section{Introduction}

\begin{figure}[!ht]
    \center
    \includegraphics[width=0.38\textwidth]
    {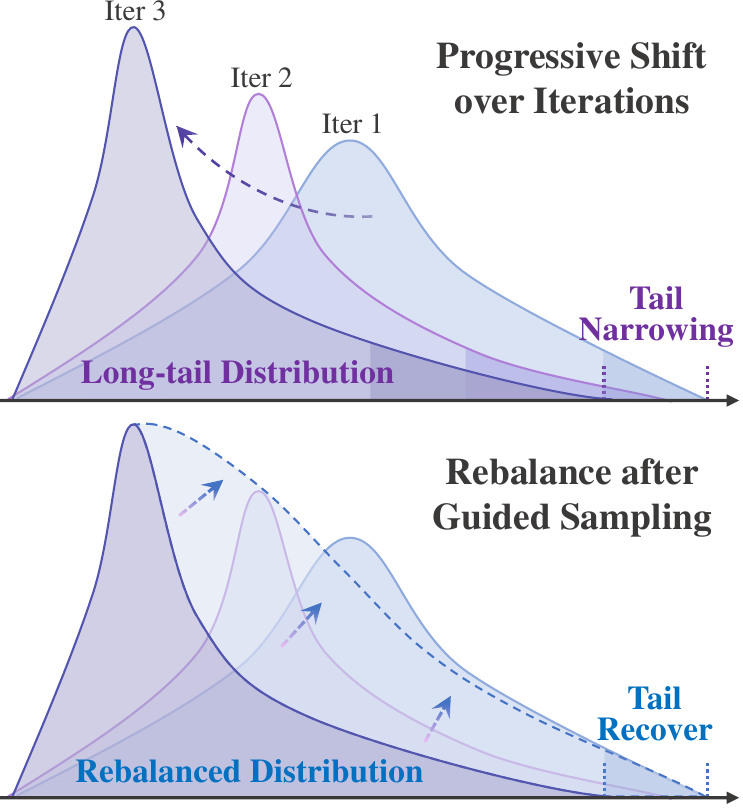} 
    \caption{Illustration of distribution during the self-improvement sampling process. \textbf{Top:} The long-tail effect intensifies with iterative training on self-generated data. The low-probability data begins to diminish, leading to \textit{tail narrowing}. \textbf{Bottem:} Guided sampling balances the distribution by improving tail data sampling efficiency.}
    \label{fig:distribution}
    \vspace{-0.5em}
\end{figure}

Large language models (LLMs) have demonstrated impressive ability in performing complex reasoning tasks \citep{wei2022chain, zerocot, llm_survey}. While fine-tuning models on curated data can further boost performance, it relies heavily on human supervision, limiting scalability and generalization \citep{gsm8k}. 
To address this, the ``self-improvement'' paradigm emerges, where models generate multiple reasoning paths, filter out incorrect responses and fine-tune themselves on their own outputs without human intervention \citep{star, rest, llm_self_improve, rest_em, self_rewarding}.

\begin{figure*}[t]
    \centering
    \includegraphics[width=\textwidth]{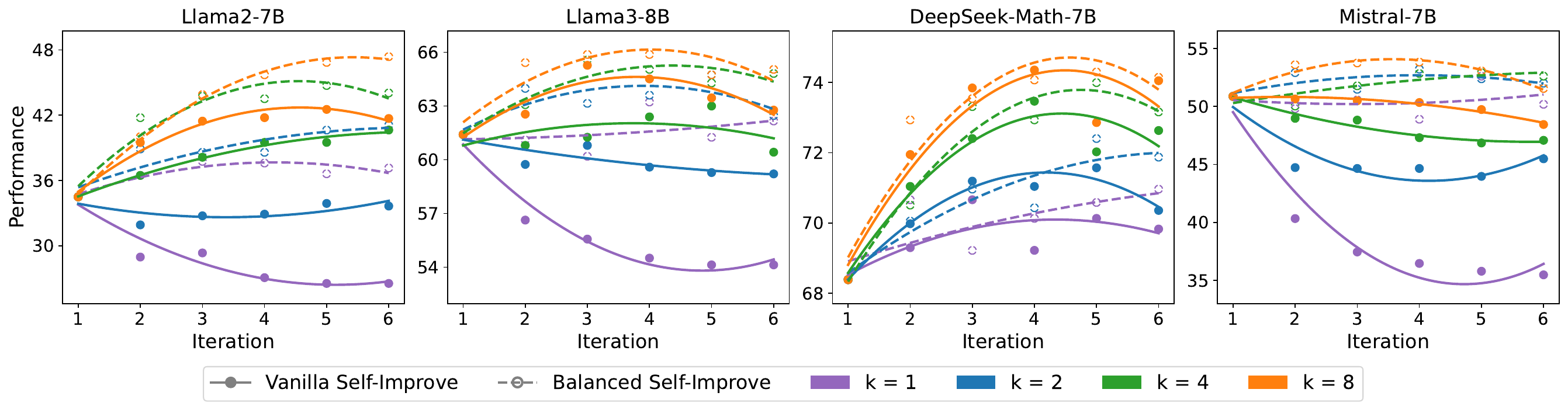}
    \caption{
    \textbf{Iterative performance in the self-improvement.} Experiments are conducted on GSM8K with varying sampling numbers $k$. 
    Solid markers show the performance of vanilla self-improve,  with the solid line fitting these points. The performance plateaus after a few iterations.
    Hollow markers represent the performance after supplementing tail data, with a dashed line trend. It balances the distribution and alleviates performance bottlenecks.}
    \label{fig:performance}
\end{figure*}

Despite the benefits of self-improvement, its performance typically reaches a ceiling after a few iterations \citep{progress_or_regress}. We perform preliminary experiments (\textsection~\ref{sec:performance_bottleneck}) and find that in reasoning tasks, the most significant gains from self-improvement occur in the first iteration, while subsequent iterations encounter performance bottlenecks or even degradation (Figure \ref{fig:performance}). 
Similar performance bottlenecks in synthetic data have also been observed in text generation \citep{curse_recursion} and image synthesis \citep{self_consuming}.

Further, we delve into the self-improvement process and conduct an in-depth analysis (Figure \ref{fig:distribution_comparison}) to investigate the underlying causes behind the performance bottlenecks. 
On the one hand, complex problems with lengthy reasoning chains tend to amplify hallucinations, making it difficult for models to explore the vast search space and sample correct rationales \citep{verify, zhang23hallucination, xie23self, r3}.
Consequently, the models tend to over-sample easy queries and under-sample queries they have yet to master \citep{dart_math}.
On the other hand, as iterations proceed, this imbalance in sampling is exacerbated, leading to a long-tail distribution where solutions to difficult queries almost disappear (Figure \ref{fig:distribution}). This situation is also referred to as \textit{tail narrowing} or \textit{tail cutting} in previous studies \citep{tale_of_tails, curse_recursion}. 
As a result, the model's self-improvement is limited, since difficult examples are also crucial for further training \citep{deita}.

To address this imbalance, a common approach is to allocate more sampling trials to under-sampled, challenging queries \citep{dart_math}, but this can be considerably more costly.
In this paper, we propose \textbf{G}uided \textbf{S}elf-\textbf{I}mprovement (\textbf{\ours}), an efficient method that leverages interactive guidance signals inspired by the Socratic method \citep{socratic, p_np}.
Specifically, our approach introduces an additional resampling phase, termed \textit{distribution re-balancing} for difficult queries, which is applied after the generation step in the self-improvement process.  
During this phase, we incorporate targeted guidance signals derived from oracle answers, stepwise rationales, and strong teacher supervision. These signals help to narrow the sampling space, reduce sampling difficulty, and minimize hallucinations during reasoning \citep{xie23self, r3}. As a result, \ours enables more effective exploration, increases solution coverage for challenging queries \citep{compute_sample}, and mitigates the issue of tail narrowing during the sampling process.

We perform experiments across four models and six mathematical reasoning tasks, including arithmetic reasoning, abstract algebra, and formal logic. The results demonstrate that \ours mitigates the performance bottlenecks of self-improvement while maintaining computational efficiency.  Further analysis shows that this method leads to a more balanced solution distribution and improved model generalization across multiple reasoning tasks. In addition to natural language reasoning, our method has also been proven effective in program-based reasoning \citep{pot}.

Our contributions are summarized as follows:
\begin{itemize}[leftmargin=*]
\item We conduct an in-depth study of the self-improvement process, revealing the performance bottlenecks driven by a long-tail distribution of solutions, which results from increasingly imbalanced data sampling.
\item To efficiently mitigate the problem of tail narrowing, we introduce the Guided Self-Improvement (\ours) method that employs Socratic-style guidance signals to assist models in exploring solutions for challenging queries.
\item We validate our strategy through comprehensive experiments on four backbone models across six mathematical reasoning tasks, demonstrating the effectiveness and efficiency of \ours.
\end{itemize}

\section{Related Work}

\subsection{Self-improvement for LLMs}
Self-improvement methods, where models refine themselves using self-generated data, have proven effective in enhancing problem-solving abilities without human intervention \citep{llm_self_improve, star}. 
To ensure the reliability of this process, the generated data is typically filtered using external supervision signals. 
These signals can be binary rewards, such as correctness checks based on reference answers \citep{rft, star, dart_math} or compiler execution feedback \citep{program_better}. 
Alternatively, more nuanced approaches involve scoring \citep{rest} or ranking systems \citep{raft}, which may be generated by the model itself \citep{self_rewarding} or external reward models \citep{vstar, rstar}. 
Some methods adopt weaker supervision signals, such as majority voting across multiple outputs \citep{llm_self_improve}.

Once filtered, the high-quality data supports post-training through methods like SFT \citep{star, isheep} or preference-based techniques like Direct Preference Optimization (DPO, \citealp{self_rewarding}). 
This process is often iterative, allowing models to continually generate new data, filter it, and use it to refine their performance further \citep{star, rest, self_rewarding}.

\subsection{Distribution Shift in Synthetic Data}
The scaling law reveals a predictable increase in model performance as the volume of training data grows \citep{scaling_law}. 
With the development of LLMs, the demand for vast amounts of high-quality data has surged, leading researchers to rely increasingly on synthetic data. 
These methods have proven effective across various tasks, from general-purpose chatbots \citep{llama3, nemotron} to specialized fields such as mathematical reasoning \citep{mammoth, metamath}.

However, the synthetic data also introduces the risk of \textit{model collapse}, where their performance degrades due to recursive training on model-generated data \citep{curse_recursion, self_consuming}. 
This phenomenon arises from \textit{distribution shift}: as models favor high-probability outputs, their results become increasingly uniform, leading to reduced variance. 
Over time, this shift manifests in the declining diversity of model responses \citep{Linguistic_Diversity}, the disappearance of tail behaviors \citep{tale_of_tails}, and the amplification of systematic bias \citep{tale_of_diversity_and_bias}.
\citet{progress_or_regress} have also observed similar degradation in self-improvement loops, which aligns with the core argument of this paper.

\begin{figure*}[t]
    \centering
    \begin{subfigure}[b]{0.32\textwidth}
        \centering
        \includegraphics[width=0.85\textwidth]{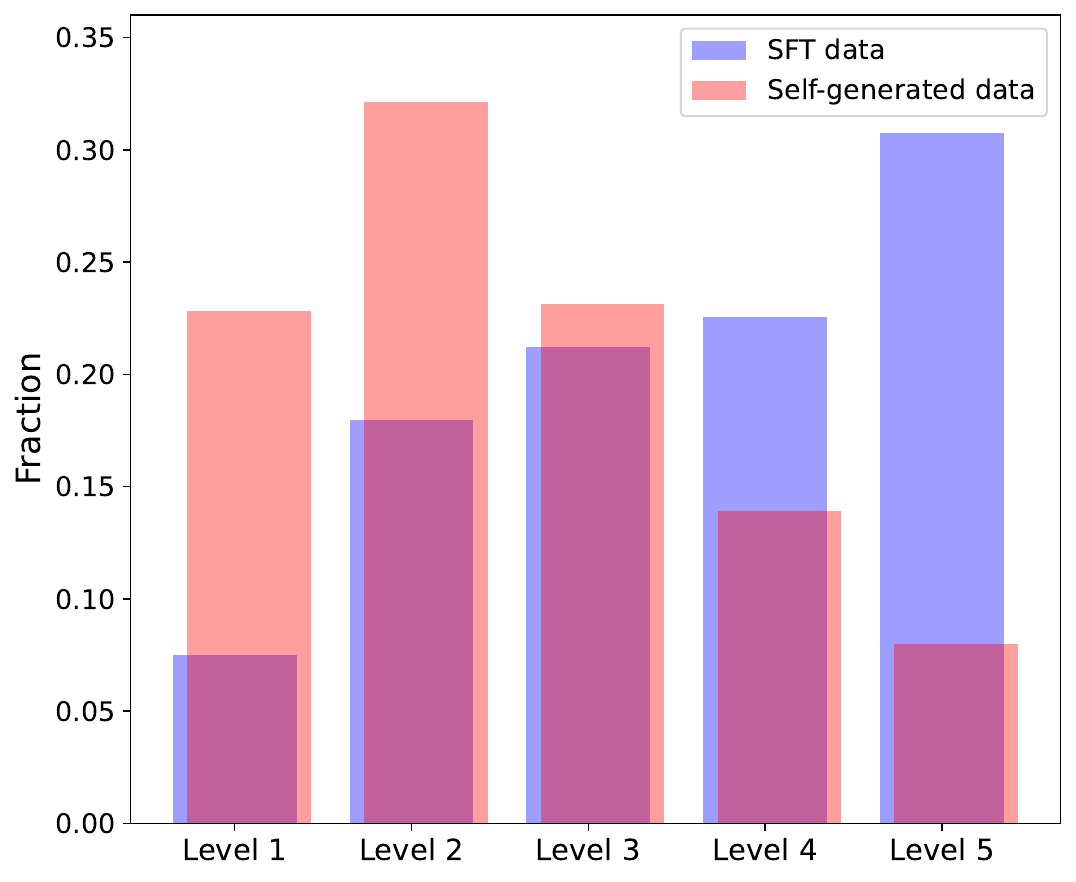}
        \caption{Difficulty distribution}
        \label{fig:difficulty}
    \end{subfigure}
    \hfill
    \begin{subfigure}[b]{0.3\textwidth}
        \centering
        \includegraphics[width=\textwidth]{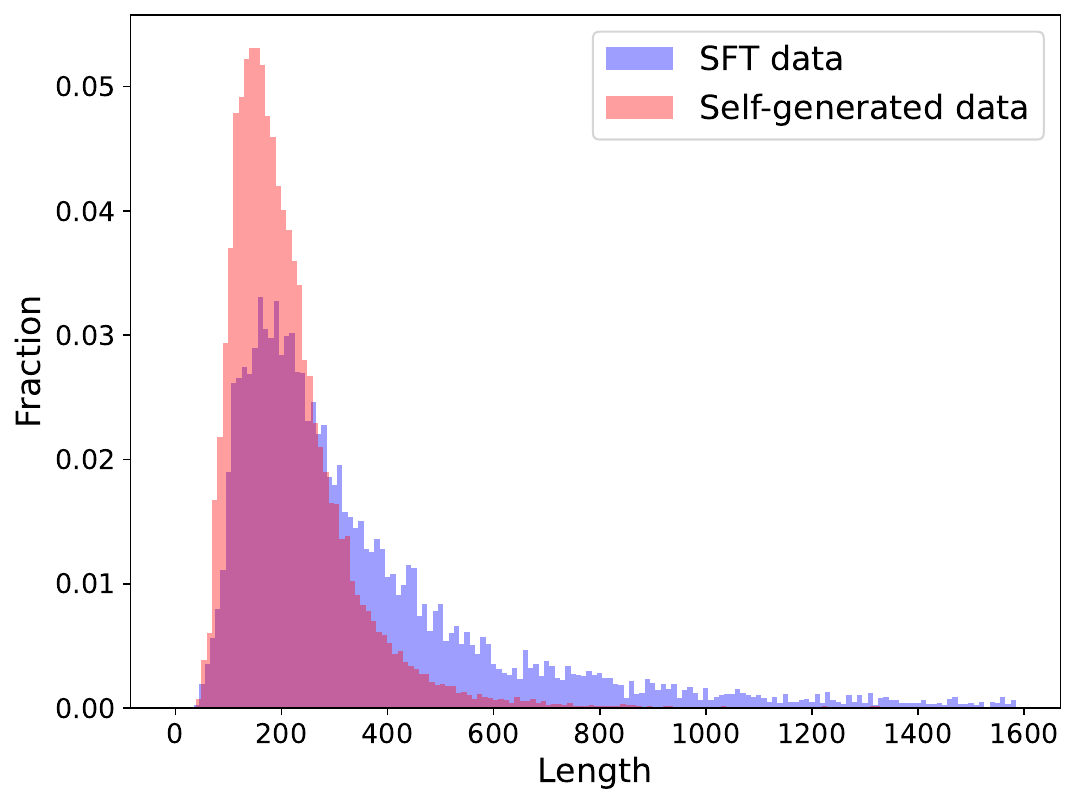}
        \caption{Length distribution}
        \label{fig:length}
    \end{subfigure}
    \hfill
    \begin{subfigure}[b]{0.3\textwidth}
    \centering\includegraphics[width=\textwidth]{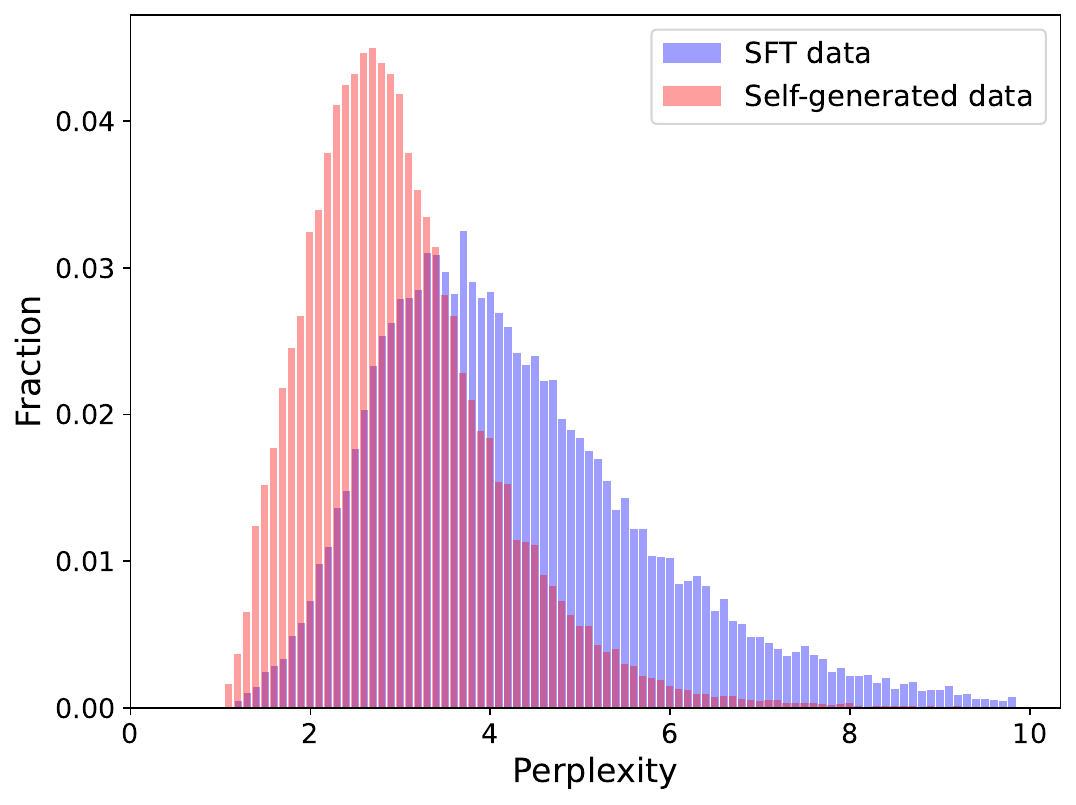}
        \caption{Perplexity distribution}
        \label{fig:ppls}
    \end{subfigure}
    \caption{\textbf{Comparison of data distributions} between the self-generated and original (SFT) datasets. \textbf{(a)} Difficulty distribution across five levels in MATH tasks, with level 1 representing the easiest and level 5 the most difficult. The self-generated data has a lower proportion of difficult problems. \textbf{(b)} Length distribution indicates that the self-generated data tends to be shorter compared to the original dataset. \textbf{(c)} Perplexity diagram of each training sequence measured with the Llama3-8B, shows that the tails in the self-generated data are diminished.}
    \label{fig:distribution_comparison}
\end{figure*}

\section{Preliminaries}

\subsection{Formulation of Self-improvement}
Given a large language model $M_0$ and the original training dataset $\mathcal{D} = \{(x_i, r_i, y_i)\}_{i=1}^{N}$, where $x_i$ is the problem, $r_i$ is the chain-of-thought rationale \citep{wei2022chain} and $y_i$ represents the final answer. Each rationale $r_i$ consists of several intermediate steps, i.e., $r_i = [r_{i,1}, \dots, r_{i,L}]$, where $L$ denotes the number of steps.
The self-improvement process enhances the model's reasoning ability through iterative refinement over $T$ cycles. Each iteration $t \in [1,T]$ consists of two main steps: \textit{Generate} and \textit{Improve}.

\paragraph{Generate step.}
At iteration $t$, the previous model $M_{t-1}$ generates multiple reasoning paths for each problem. Specifically, we allocate $k$ sampling times to each query $x_i \in \mathcal{D}$:
\begin{equation}
(\hat{r}_i, \hat{y}_i) = M_{t-1}(x_i) \nonumber
\end{equation}
The newly generated data points form a set $\mathcal{D}^{'}=\{(x_i, \hat{r}_i^j, \hat{y}_i^j) \mid x_i \in \mathcal{D}, j=[1,k]\}$. Each candidate solution $\hat{y}_i^j$ is evaluated by a binary reward function $\textrm{rf} (y_i, \hat{y_i}) \in \{0, 1\}$, which verifies correctness based on ground-truth answers $y_i$. Only correct solutions with $\textrm{rf}(\cdot) = 1$ are filtered to form the high-quality dataset $\mathcal{D}_t$ for the current iteration $t$.

\paragraph{Improve step.}
In the $t$-th iteration, the model $M_t$ is fine-tuned on the self-generated high-quality dataset $\mathcal{D}_t$.
The fine-tuning objective is to minimize the negative log-likelihood (NLL) loss: 
\begin{equation}
\mathcal{L}_{\text{SFT}} = -\mathbb{E}_{(x,r)\sim \mathcal{D}_{t}} \sum\limits_{l=1}^{L} \log M (r_l \mid r_{<l}, x). \nonumber
\end{equation}

By minimizing $\mathcal{L}_{\text{SFT}}$, the model iteratively improves its ability to generate correct rationales, and this process is repeated for $T$ iterations to achieve self-improvement. Note that in the first iteration, we directly fine-tune $M_0$ on the original dataset $\mathcal{D}$ to obtain $M_1$.

\subsection{Biased Sampling and Tail Effect}
In the self-improvement sampling process, there is a tendency to select higher-quality and more accurate data, which introduces a phenomenon known as sampling bias \citep{self_consuming, curse_recursion}. This bias often results in the truncation or narrowing of low-probability ``tails'' in the data distribution.

To illustrate the effects of sampling bias, consider a one-dimensional Gaussian distribution \( X^0 \sim \mathcal{N}(\mu, \sigma^2) \). Let \( \lambda \in [0, 1] \) represent the sampling bias parameter and biased sampling from \( \mathcal{N}(\mu, \lambda \sigma^2) \). When \( \lambda = 1 \), the sampling is unbiased, maintaining the original variance. Conversely, when \( \lambda = 0 \), the sampling is derived from the modes of the generative distribution $M_t$ with zero variance.

Initially, the tails of the distribution, which represent low-probability data, begin to diminish due to their low likelihood of being sampled. As the iteration continues, the heavy-tailed data is increasingly excluded, causing a shift in the overall distribution. This phenomenon is referred to as tail narrowing or tail cutting, leading to a more peaked distribution \citep{tale_of_tails}.
\section{Performance Bottleneck and Tail Narrowing in Self-Improvement}
\label{sec:performance_bottleneck}
Despite extensive research into self-improvement methods \citep{rest, rest_em}, the performance dynamics across successive training iterations remain underexplored.
To this end, we conduct experiments across four backbone models to investigate the effects of sampling and tail data during iterative training.
Tail data refers to samples for which the model rarely generates correct solutions during sampling. These samples lie in the ``long tail" of data distributions.
\paragraph{Performance trends.}
To uncover the relationship between performance trends with sampling, we vary the number of sampling $k$.
Figure~\ref{fig:performance} reveals the following:
(1) Lower sampling times (e.g., $k=1$ or $k=2$) degrade model performance, leading to negative gains. Llama2-7B, Mistral-7B, and Llama3-8B underperform compared to SFT on the original dataset (iteration 1). This decline stems from the models' weaker reasoning abilities, which limit their coverage of challenging queries.
Consequently, the models tackle only basic problems, resulting in a degradation or collapse of reasoning ability \citep{curse_recursion, strong_collapse}.
(2) With larger sampling numbers $k$, performance improves across multiple iterations. The most notable gains occur during the first iteration, as shown by the green and orange solid lines in Figure~\ref{fig:performance}. However, after the third iteration, progress halts, eventually reaching a performance plateau.

\paragraph{Impact of tail data.}
As previously discussed, merely scaling the number of sampling eventually encounters performance bottlenecks. 
However, challenging low-probability samples, often referred to as tail data, are often considered more crucial for improving model performance \citep{beyond_neural_scaling, deita, dart_math}.

To investigate the impact of tail data, we conduct additional experiments targeting these difficult examples.
For queries that do not yield a correct response after $k$ sampling attempts, we supplement them with a golden rationale, ensuring that every query has at least one practically correct response and effectively rebalancing the long-tail distribution.
We use hollow markers to represent the performance with tail data.
As shown in Figure \ref{fig:performance}, across different numbers of sampling and model variants, the performance represented by the dashed line exceeds that of the solid line representing vanilla self-improve.
The results demonstrate that rebalanced data helps mitigate performance degradation under a small number of sampling attempts (e.g., $k=1$ or $k=2$).
Moreover, as $k$ increases, a larger number of sampling alleviates performance plateaus and boosts model efficacy to some extent.
This study highlights the value of solutions to challenging queries in overcoming the limits of finite and progressively biased sampling.

\paragraph{Emergence of tail narrowing.}
To analyze the distributional characteristics of tail data, we analyze three dimensions: difficulty, response length, and perplexity, revealing the differences between self-generated and original data.
Figure \ref{fig:difficulty} shows that model-generated data tends to be at a low level and simpler compared to the original MATH dataset, which contains more challenging queries. 
Consistently, DART-Math \citep{dart_math} also identifies a synthesis bias towards easy queries.

Using response length as a complexity indicator, as suggested by \citet{complexity_cot}, we find that self-generated responses are generally shorter, peaking around $200$ tokens (Figure \ref{fig:length}). This indicates a bias towards simplicity and a lower proportion of complex data.

From a semantic perspective, we computed the perplexity based on  Llama3-8B. As shown in Figure \ref{fig:ppls}, the self-generated data has lower perplexity, indicating a shift toward more probable and coherent tokens. This reduces the occurrence of rare, complex and diverse tokens in the tail.

These observations indicate a diminishing trend of tail data, termed \textit{Tail Narrowing} \citep{tale_of_tails}.
In each iteration, the distribution progressively narrows, hindering performance to challenging queries.

\section{Guided Self-Improvement}

To mitigate the observed tail-narrowing phenomenon, a straightforward solution is to increase the sampling trials for tail data.
However, directly tackling these difficult queries from scratch often results in low success rates and higher costs due to repeated failed attempts.
Drawing from Socratic-style education \citep{socratic, p_np}, we introduce guidance-based exploration techniques to improve sampling efficiency, such as learning from demonstrations \citep{learning_from_demos, exploration_from_demos}.
We provide the model with tailored assistance in structured contexts, enabling it to address difficult queries more effectively.
The following paragraphs outline four guiding strategies we propose.

\paragraph{Answer-driven.}
This strategy incorporates the ground-truth answer $y_i$ along with the input query $x_i$ as context to guide the generation process.
It helps the model better align with the expected solution, particularly for challenging queries \citep{star}.
Formally, at each iteration $t$, instead of inputting only $x_i$ into the model $M_{t-1}$, we extend the prompt by appending $y_i$ as a hint:
\begin{equation}
(\hat{r}_i, \hat{y}_i)  = M_{t-1}(x_i, \text{hint}(y_i)). \nonumber
\end{equation}
This approach helps the model focus on the reasoning process behind the answer, reducing the overall difficulty of the task.

\paragraph{Rationale-driven.}
In this strategy, we further extend the input by introducing a rationale $r_i$, which helps the model derive the correct reasoning process.
Unlike the answer-driven method, providing a rationale offers a more detailed reference for the model to follow, narrowing the exploration space of reasoning paths \citep{sdft}.
Formally, at iteration $t$, the input to the model $M_{t-1}$ is augmented as follows:
\begin{equation}
(\hat{r}_i, \hat{y}_i)  = M_{t-1}(x_i, \text{hint}(r_i)). \nonumber
\end{equation}
This approach enables the model to handle queries it has yet to master and ensures a higher coverage of solved problems.
Importantly, it alleviates the hallucinations when the model tries to provide reasoning paths for problems it doesn't fully understand \citep{cot_faithfulness}, improving the reliability of generated data.

\paragraph{Interactive sampling.}
Inspired by previous work in the area of Interactive RL \citep{exploration_from_demos, suay2011effect}, we introduce feedback from a stronger model $M_s$ after the model $M_{t-1}$ fails. 
Instead of providing hints along with the query, this dynamic process ensures that the model can explore its own solution before receiving external guidance.
Formally, after the model $M_{t-1}$ generates an incorrect answer, we re-sample by giving it both its prior incorrect output and the feedback $f_i$ from $M_s$ as additional context:
\begin{flalign*}
&(\hat{r}_i^{\mathrm{error}}, \hat{y}_i^{\mathrm{error}}) = M_{t-1}(x_i), \\
&f_i = M_s(x_i, r_i, y_i, \hat{r}_i^{\mathrm{error}}), \\
&(\hat{r}_i, \hat{y}_i) = M_{t-1}(x_i, \hat{r}_i^{\mathrm{error}}, f_i).
\end{flalign*}
This feedback includes an analysis and correction of the model's errors, reducing its reliance on the correct answer. 
Through this interactive process, we balance exploration and correction, enabling the model to learn from its mistakes without overly restricting its reasoning path.

\paragraph{State reset.}
Drawing inspiration from the concept of \textit{state reset}, i.e., going back to intermediate states during problem-solving to refine the approach \citep{dataset_reset, r3}, we adopt a strategy where the model is guided step by step with partial rationales. 
Instead of supplying the full rationale $r_i$ immediately, after $l$ incorrect attempts, the model is provided with the preceding reasoning steps $r_{<l}$ from $r_i = [r_{i,1}, \dots, r_{i,L}]$, gradually narrowing down the exploration space:
\begin{equation}
(\hat{r}_i, \hat{y}_i) = M_{t-1}(x_i, \text{hint}(r_{i,<l})). \nonumber
\end{equation}
This method reduces the difficulty of queries at a fine-grained level, relieving the model of cognitive overload while still allowing flexibility in the model’s solution. 
Although it increases the number of attempts, the incremental hint helps identify the threshold where the model can solve problems independently with minimal guidance.

\section{Experiments}

\begin{table*}[t]
\belowrulesep=1pt
\aboverulesep=1pt
\centering
\setlength{\tabcolsep}{3pt}
\resizebox{\textwidth}{!}{%
\begin{tabular}{llcccccccccc}
\toprule
\multirow{2}{*}[-1pt]{\textbf{Models}} & \multirow{2}{*}[-1pt]{\textbf{Methods}}      & \multirow{2}{*}[-2pt]{\makecell{\textbf{Sample}\\\textbf{Budget}}}   & \multirow{2}{*}[-1pt]{\textbf{Coverage}}  & \multicolumn{4}{c}{\textbf{Held-in Datasets}} & \multicolumn{4}{c}{\textbf{Held-out Datasets}} \\
\cmidrule(r){5-8} \cmidrule(r){9-12}
           &      &      &     & \myblue\textbf{Avg.} & \textbf{AQuA}   & \textbf{GSM8K}  & \textbf{MATH}   & \myblue\textbf{Avg.} &  \textbf{MathQA} & \textbf{SVAMP}  & \textbf{Thm.QA}        \\
\midrule
\multirow{9}{*}{\textbf{Llama2-7B}}  & SFT   & - & - & \myblue$21.00$  & $27.17$  & $31.31$ & $4.52$   & \myblue$21.08$  & $22.08$   & $36.80$  & $4.38$ \\
           & SI ($k=8$)   & $0.24\text{M}$ & $52.6\%$ & \myblue$21.53$  & $25.59$  & $33.13$ & $5.86$   & \myblue$24.46$  & $27.37$   & $39.50$  & $6.50$  \\
           & Brute-Force SI ($k=64$)  & $1.11\text{M}$ & $77.1\%$ & \myblue$23.73$  & $29.53$  & $35.86$ & $5.80$  & \myblue$23.42$  & $25.76$   & $39.50$  & $5.00$ \\
           & Brute-Force SI ($k=128$) & $1.67\text{M}$ & $80.7\%$ & \myblue$23.84$  & $27.17$  & \underline{$37.98$} & $6.36$   & \myblue$24.22$  & $26.57$   & $40.60$  & $5.50$ \\
           \cdashline{2-12}\cdashline{2-12}
           & Guided Self-Improve ($k=8$) &       &       &          &           &       &           &        &          &                    &        \\
           & \ \ + Answer-driven        & $0.37\text{M}$ & $99.0\%$ & \myblue$23.44$  & $28.34$  & $35.94$ & $6.04$  & \myblue$25.27$ & $25.83$   & $\mathbf{43.60}$  & $6.38$ \\
           & \ \ + Rationale-driven     & $0.34\text{M}$ & $99.9\%$ & \myblue$24.32$  & $30.32$  & $36.16$ & $\mathbf{6.48}$ & \myblue$25.68$   & \underline{$28.11$}   & $41.30$  & $\mathbf{7.63}$ \\
           & \ \ + Interactive Sampling & $0.36\text{M}$ & $96.3\%$ & \myblue\underline{$25.00$} & \underline{$30.71$} & $37.83$ & $\underline{6.46}$  & \myblue\underline{$26.25$}  & $27.84$   & \underline{$43.40$}  & \underline{$7.50$} \\
           & \ \ + State Reset          & $0.38\text{M}$ & $82.0\%$ & \myblue$\mathbf{25.91}$ & $\mathbf{31.10}$ & $\mathbf{40.18}$ & $6.44$  & \myblue$\mathbf{26.79}$  & $\mathbf{29.45}$   & $43.30$  & $\mathbf{7.63}$ \\
\midrule
\multirow{9}{*}{\textbf{Llama3-8B}}     & SFT   & - & - & \myblue$37.27$ & $39.37$  & $57.47$ & $14.96$ & \myblue$38.68$  & $44.29$   & $63.00$  & $8.75$ \\
           & SI ($k=8$)       & $0.24\text{M}$ & $68.6\%$ & \myblue$36.87$ & $39.76$  & $59.14$ & $11.70$  & \myblue$39.09$  & $45.63$   & $62.40$  & $9.25$ \\
           & Brute-Force SI ($k=64$)      & $0.85\text{M}$ & $86.4\%$ & \myblue$38.16$ & $38.98$  & $61.11$ & $14.40$ & \myblue$38.30$  & $45.86$   & $60.90$  & $8.13$ \\
           & Brute-Force SI ($k=128$)     & $1.26\text{M}$ & $88.7\%$ & \myblue$37.55$  & $41.34$  & \underline{$61.64$} & $9.68$   & \myblue$39.32$ & $45.90$   & $62.80$  & $9.25$ \\
           \cdashline{2-12} \cdashline{2-12}
           & Guided Self-Improve ($k=8$) &       &       &          &           &       &           &        &          &                    &        \\
           & \ \ + Answer-driven        & $0.31\text{M}$ & $98.5\%$ & \myblue$38.71$ & $42.52$  & $59.82$ & $13.80$ & \myblue$40.15$  & $46.85$   & $63.60$  & \underline{$10.00$} \\
           &  \ \ + Rationale-driven    & $0.29\text{M}$ & $99.8\%$ & \myblue$39.14$ & \underline{$42.91$}  & $60.27$ & $14.24$ & \myblue$41.08$  & \underline{$47.94$}   & $\mathbf{65.30}$  & \underline{$10.00$} \\
           & \ \ + Interactive Sampling & $0.31\text{M}$ & $97.3\%$ & \myblue\underline{$39.34$} & $42.52$  & $60.05$ & \underline{$15.46$} & \myblue\underline{$41.12$}   & $47.24$   & $64.50$  & $\mathbf{11.63}$ \\
           & \ \ + State Reset          & $0.32\text{M}$ & $90.2\%$ & \myblue$\mathbf{41.64}$ & $\mathbf{46.46}$  & $\mathbf{62.62}$ & $\mathbf{15.54}$ & \myblue$\mathbf{41.33}$  & $\mathbf{49.25}$   & \underline{$65.00$}  & $9.75$ \\
\midrule
\multirow{9}{*}{\makecell{\textbf{DeepSeek-}\\\textbf{Math-7B}}} & SFT   & - & - & \myblue$50.01$ & $60.63$  & $60.73$ & $28.66$ & \myblue$21.15$  & $21.61$   & $37.10$  & $4.75$ \\
           & SI ($k=8$)   & $0.24\text{M}$ & $79.8\%$ & \myblue$52.22$ & $56.69$  & $68.92$ & $31.06$ & \myblue$49.63$  & $64.26$   & $67.50$  & $17.13$ \\
           & Brute-Force SI ($k=64$)   & $0.64\text{M}$ & $91.5\%$ & \myblue$53.95$ & $57.87$  & $70.89$ & $\mathbf{33.08}$ & \myblue$50.92$  & \underline{$65.36$}   & $70.40$  & $17.00$ \\
           & Brute-Force SI ($k=128$)     & $0.93\text{M}$ & $92.2\%$ & \myblue$52.43$  & $59.45$  & $72.02$ & $25.82$   & \myblue$48.76$ & $64.69$   & $68.20$  & $13.38$ \\
           \cdashline{2-12}\cdashline{2-12}
           & Guided Self-Improve ($k=8$) &       &       &          &           &       &           &        &          &                    &        \\
           & \ \ + Answer-driven        & $0.30\text{M}$ & $99.5\%$ & \myblue$53.83$ & \underline{$61.02$}  & $70.05$ & $30.42$ & \myblue$51.80$  & $64.32$   & $\mathbf{72.70}$  & $\mathbf{18.38}$ \\
           &  \ \ + Rationale-driven    & $0.29\text{M}$ & $99.7\%$ & \myblue$53.71$ & $58.66$  & $71.65$ & $30.82$ & \myblue$51.20$  & $64.02$   & $72.20$  & $17.38$ \\
           & \ \ + Interactive Sampling & $0.30\text{M}$ & $96.9\%$ & \myblue$\mathbf{55.67}$ & $\mathbf{61.81}$  & \underline{$72.63$} & $32.56$ & \myblue\underline{$51.69$}  & $\mathbf{66.83}$   & $71.10$  & $17.13$ \\
           & \ \ + State Reset          & $0.31\text{M}$ & $93.6\%$ & \myblue\underline{$55.04$} & $59.45$  & $\mathbf{72.78}$ & \underline{$32.90$} & \myblue$\mathbf{51.85}$  & $64.59$   & $\mathbf{72.70}$  & \underline{$18.25$} \\
\midrule
\multirow{9}{*}{\textbf{Mistral-7B}} & SFT   & - & - & \myblue$28.27$ & $31.10$  & $44.96$ & $8.74$ & \myblue$21.08$  & $22.08$   & $36.80$  & $4.38$ \\
           & SI ($k=8$)   & $0.24\text{M}$ & $61.9\%$ & \myblue$25.22$ & $27.95$  & $40.56$ & $7.14$  & \myblue$26.96$  & $33.84$   & $43.30$  & $3.75$ \\
           & Brute-Force SI ($k=64$)   & $0.87\text{M}$ & $82.5\%$ & \myblue$28.23$ & $28.74$  & \underline{$47.16$} & $8.80$  & \myblue$30.16$  & $34.94$   & $49.90$  & $5.36$ \\
           & Brute-Force SI ($k=128$)     & $1.27\text{M}$ & $85.1\%$ & \myblue$28.32$  & $30.32$  & $45.79$ & $8.84$   & \myblue$28.81$ & $33.03$   & $49.40$  & $4.00$ \\
           \cdashline{2-12}\cdashline{2-12}
           & Guided Self-Improve ($k=8$) &       &       &          &           &       &           &        &          &                    &        \\
           & \ \ + Answer-driven        & $0.33\text{M}$ & $98.3\%$ & \myblue$28.09$ & $\mathbf{34.65}$  & $42.00$ & $7.62$  & \myblue$29.70$  & $34.34$   & $49.90$  & $4.88$ \\
           &  \ \ + Rationale-driven    & $0.31\text{M}$ & $99.7\%$ & \myblue\underline{$29.13$} & $32.68$  & $46.17$ & $8.54$ & \myblue$\mathbf{32.14}$  & $\mathbf{35.24}$   & $\mathbf{53.80}$  & $\mathbf{7.38}$ \\
           & \ \ + Interactive Sampling & $0.31\text{M}$ & $96.4\%$ & \myblue$29.04$ & $32.28$  & $45.87$ & \underline{$8.98$} & \myblue\underline{$30.22$}  & \underline{$35.04$}   & $50.00$  & \underline{$5.63$} \\
           & \ \ + State Reset          & $0.34\text{M}$ & $86.7\%$ & \myblue$\mathbf{31.23}$ & \underline{$33.07$} & $\mathbf{50.95}$ & $\mathbf{9.68}$  & \myblue$30.21$  & $34.94$   & \underline{$50.20$}  & $5.50$  \\
\bottomrule
\end{tabular}
}
\caption{Main results on six math reasoning tasks. The best result for each dataset is highlighted in \textbf{bold}, while the second-best result is marked with \underline{underline}.
Results marked in \colorbox{lightblue}{blue} indicate average scores.
Thm.QA denotes the TheoremQA task.
\textit{Coverage} refers to the number of unique problems solved in the Generate Step, while \textit{Sample Budget} indicates the total number of sampling times during this step.
``SI'' refers to Self-Improve.
The baselines include SFT, the vanilla Self-Improve, and Brute-Force Self-Improve.
} 
\label{tab:main_result}
\end{table*}

\subsection{Experimental Setups} 

\paragraph{Models.}
We conduct experiments using four widely adopted foundation models, including Llama2-7B-Base \citep{llama2}, Llama3-8B-Base \citep{llama3}, Deepseek-Math-7B-Base \citep{deepseek}, and Mistral-7B-v0.3 \citep{mistral}. For the stronger model in the interactive sampling process, we employ Llama3-70B-Instruct \citep{llama3}.

\paragraph{Datasets.}
We utilize six math reasoning datasets. These include arithmetic reasoning datasets such as GSM8K \citep{gsm8k}, AQuA \citep{AQuA}, MathQA \citep{mathqa} and SVAMP \citep{svamp}, as well as a more challenging dataset MATH \citep{MATH}.
We also include TheoremQA for abstract algebra and formal logic.
To evaluate generalization, we choose AQuA, GSM8K, MATH as held-in datasets and MathQA, SVAMP, TheoremQA as held-out datasets.
To ensure consistency in answer format across different datasets, we utilize the unified data provided by the MathInstruct dataset \citep{mammoth} and follow its train-test splits. More dataset statistics can be found in Appendix \ref{app:data}.

\paragraph{Implementation details.}
Following \citet{llm_self_improve} and \citet{rest_em}, we fine-tune the pre-trained model $M_0$ during the Improve Step of each iteration to prevent overfitting.
We set the iteration number $T=4$ and sampling number $k=8$.
To mitigate the tail-narrowing effect, we identify queries with less than a 50\% probability of yielding correct completions as heavy-tailed data. For the tail data, we apply the \ours strategy, resampling up to $k$ times until the query no longer falls into the tail data.
All experiments are performed on 8 A100-80GB GPUs.
We run the SFT and Improve Step for $1$ epoch. 
The learning rate is set to $1 \times 10^{-5}$.
For sampling and evaluation, we leverage the vLLM \citep{vllm} framework, setting a maximum of $1024$ output tokens.  
The temperature is set to $0.7$ during sampling and $0$ during evaluation.
The prompt templates are detailed in Appendix \ref{app:prompt}.

\paragraph{Baselines.}
To evaluate the impact of \ours on distributional adjustments, we compare it against SFT and Self-improve variants. We also include baselines with significantly higher sampling trials than \ours. 
\begin{itemize}[leftmargin=*]
\item \textbf{SFT}: Fine-tuning on the original dataset for $1$ epoch, which corresponds to the first iteration of self-improvement.
\vspace{-0.5em}
\item \textbf{Self-Improve ($k=8$)}: In each iteration, we sample $k=8$ completions per query from the original dataset, filter out correct reasoning paths, and then improve on the self-generated data.
\vspace{-0.5em}
\item \textbf{Brute-Force Self-Improve} ($k=64$ or $k=128$): Based on Vanilla Self-Improve ($k=8$), we add a distribution re-balancing stage for tail data in each iteration. It performs non-guided brute-force sampling of tail-end data up to $k$ times without additional guidance.
\end{itemize}

\subsection{Main Results}
The main results are shown in Table \ref{tab:main_result}. We have the following key findings: 

\paragraph{Re-balancing tail data improves coverage and performance of self-improvement.}
Compared to SFT, vanilla self-improve with $k=8$  boosts reasoning on held-out datasets but shows only marginal gains, with performance bottlenecks and degradation on held-in datasets.
This aligns with our observation (\textsection~\ref{sec:performance_bottleneck}), where we identify tail narrowing as the primary cause.
To mitigate this, Brute-Force Self-Improve ($k=64$ or $k=128$) performs additional sampling on tail data, which re-balances the distribution.
This adjustment significantly enhances both coverage and overall performance, with further improvements as the number of samples increases.
For example, in Llama2-7B, coverage increases from $52.6$\% to $80.7$\%, with performance improving from $21.53$ to $23.84$. In Mistral-7B, the rebalancing also reverses the observed performance decline.
Thus, incorporating a resampling stage for challenging, heavy-tailed data proves essential.

\paragraph{\ours outperforms brute-force sampling with better efficiency.}
To optimize the amount of sampling computation required for self-improvement, we analyze the sampling efficiency during the Generate step phase.
For each query $x_i \in \mathcal{D}$ at iteration $t$, we perform $k_{i,t}$ sampling operations. The cumulative sample budget across all queries and $T$ iterations is defined as:
$\text{Sample Budget} = \sum_{x_i \in \mathcal{D}} \sum_{t=1}^{T} k_{i,t}$.

When scaling resampling operations from 64 to 128, we observe that the improvement in sampling coverage becomes slower, and the performance gains diminish despite the additional computational costs.
This suggests that the model may get trapped in vast search space, particularly when dealing with more challenging queries.
In contrast, our strategy \ours, which leverages Socratic-style guidance, achieves more compute-efficient sampling. 
It outperforms brute-force sampling while using only one-third of the sampling budget.
Specifically, on Llama3-8B, the state reset strategy performs $0.32$M sampling, which is only one-third of the budget required for brute-force sampling. 
Moreover, the model shows improved held-in performance from $37.55$ to $41.64$ and generalizes well to held-out datasets.

\paragraph{The effectiveness of different strategies.}
Among the four strategies, the state reset strategy, which samples from different initial states and generates diverse reasoning paths, performs relatively better.
However, the effectiveness of different strategies depends on the model's inherent capabilities.
For example, the answer-driven strategy, which provides only the correct value and requires the model to reason backward to generate a rationale, demands advanced reasoning abilities \citep{star}. 
Therefore, this approach yields modest performance gains on weaker models such as Llama2-7B. 
Further investigation is needed to explore how different models can be optimally paired with various strategies.

\section{Discussion}

\begin{figure}[t]
    \center
    \includegraphics[width=0.48\textwidth]
    {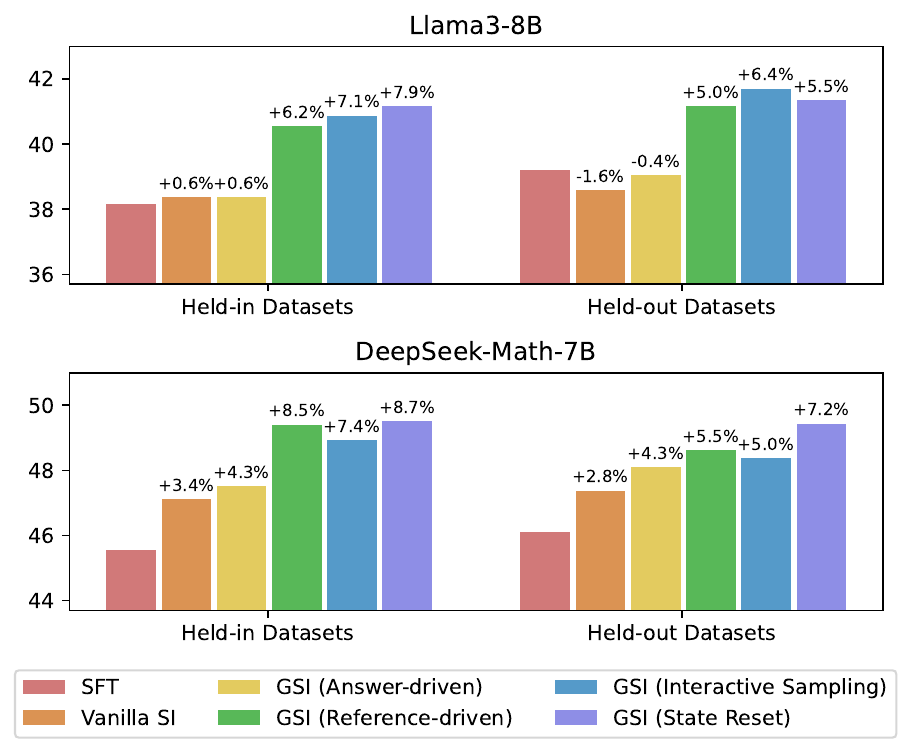} 
    \caption{Comparison of average performance on six math tasks using PoT. The percentage of improvement is significant in the held-in datasets.}
    \label{fig:pot_result}
\end{figure}

\subsection{Effectiveness on PoT Reasoning}
To fully exploit the potential of diverse reasoning processes, we extend our investigation to Program-of-Thought (PoT, \citealp{pot}) prompting.
In the self-improvement process, we utilize PoT rationales for training, then filter data and evaluate performance based on compiler-executed results.
As shown in Figure \ref{fig:pot_result}, four strategies consistently outperform the self-improvement baseline in program-based reasoning.
The state reset strategy on the DeepSeek-Math-7B model shows notable relative gains, with an improvement of up to $8.7$\%. Similarly, the reference-driven strategy leads to a performance boost of $8.5$\%.

\begin{table}[t]
\centering
\resizebox{0.46\textwidth}{!}{%
\begin{tabular}{llcc}
\toprule
\textbf{Models} & \textbf{Methods} & \textbf{Held-in} & \textbf{Held-out} \\
\midrule
\multirow{6}{*}{\makecell{\textbf{DeepSeek-}\\\textbf{Coder-1.3B}}} & SFT & $10.66$ & $14.75$ \\
 & Vanilla Self-Improve & $12.61$ & $13.98$ \\
 \cdashline{2-4}\cdashline{2-4}
 & \ours (Answer-driven) & $13.20$ & $15.62$ \\
 & \ours (Rationale-driven) & \underline{$14.99$} & $16.37$ \\
 & \ours (Interactive) & $14.13$ & \underline{$15.94$} \\
 & \ours (State Reset) & $\mathbf{16.04}$ & $\mathbf{17.38}$ \\
\midrule
\multirow{6}{*}{\makecell{\textbf{CodeLlama}\\\textbf{-13B}}}& SFT & $23.43$ & $28.15$ \\
 & Vanilla Self-Improve & $27.85$ & $29.99$ \\
 \cdashline{2-4}\cdashline{2-4}
 & \ours (Answer-driven) & $31.11$ & $31.57$ \\
 & \ours (Rationale-driven) & $31.38$ & $30.13$ \\
 & \ours (Interactive) & \underline{$32.27$} & $\mathbf{33.77}$ \\
 & \ours (State Reset) & $\mathbf{33.04}$ & \underline{$31.68$} \\
\bottomrule
\end{tabular}%
}
\caption{Effectiveness of \ours on different sizes of models. The improvement becomes more pronounced as the model size increases.}
\vspace{-1em}
\label{tab:different_model_size}
\end{table}

\subsection{Performance of Different Model Sizes}
To further explore, we investigate the effectiveness of the proposed strategy across different model sizes.
We choose a smaller model, DeepSeek-Coder-1.3B \citep{deepseek_coder}, and a larger model, CodeLlama-13B \citep{codellama}, in our experiments. As shown in Table \ref{tab:different_model_size}, DeepSeek-Coder-1.3B exhibits only marginal improvements when applying the answer-driven strategy compared to the others. 
The limited improvement may be attributed to the nature of the strategy, which requires the model to reverse-engineer a solution from a given true answer \citep{star}.
While the final result is provided, deriving a good justification can be challenging for smaller models \citep{wei2022emergent}.
However, when scaling up to the 13B model, we observe a pronounced performance boost. The results suggest that \ours is more effective with larger models, which are equipped with advanced reasoning abilities.

\subsection{Short-cutting in Generated Rationales}
\begin{figure}[t]
    \center
    \includegraphics[width=0.46\textwidth]
    {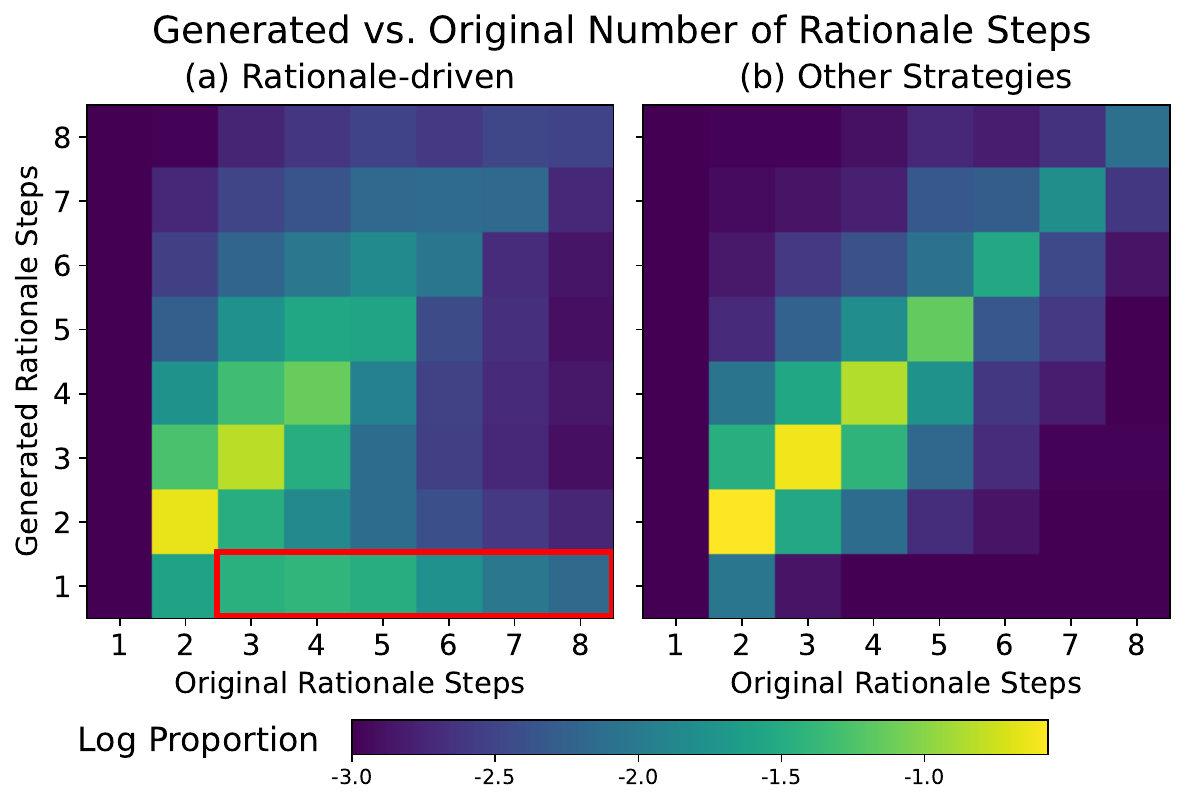} 
    \caption{Comparison of the number of rationale steps generated by the model relative to the number of steps used in the ground truth. The red box highlights the occurrence of skip steps in the rationale-driven strategy.}
    \label{fig:heatmap}
    \vspace{-1em}
\end{figure}

This experiment explores how different guiding strategies influence the quality of rationales.
Inspired by \citet{star}, we focus on the number of rationale steps.
Figure \ref{fig:heatmap} shows that, in most cases, the model's reasoning steps align with the original annotated steps. 
However, under the rationale-driven strategy, the model is more likely to generate fewer steps than others.
Further analysis reveals that providing rationales can cause the model to skip reasoning steps, a phenomenon we refer to as ``Hint Short-cutting'' \citep{star}.
This tendency may weaken the model's ability to think step-by-step, potentially hindering iterative training \citep{strong_collapse}. 
We show an example in Appendix~\ref{app:error_patterns}, where the model skips critical steps.

\subsection{Impact of the Sampling Hyperparameter}

We investigate the impact of the hyperparameter $k$, which determines the number of sampling times allocated to each query. As shown in Table \ref{tab:different_sampling_times}, increasing $k$ yields substantial performance improvements. Notably, the most significant improvements occur up to $k=8$, after which the gains begin to plateau. Therefore, we choose $k=8$  as a practical configuration, achieving a balanced trade-off between computational cost and performance gains.

\section{Conclusion}

In this work, we delve into the performance bottlenecks in the self-improvement process of LLMs, identifying the issue of tail narrowing caused by progressively imbalanced data sampling.
To mitigate this, we propose Guided Self-Improvement (\ours), a new method that incorporates a distribution re-balancing phase and Socratic-style guidance to enhance solution coverage for challenging queries. 
Experimental results across multiple models and mathematical reasoning tasks demonstrate the effectiveness of this method in improving reasoning performance while maintaining computational efficiency.
We believe \ours offers a promising direction for enhancing the scalability and generalization of self-improving models in the future.

\begin{table}[t]
  \centering
\resizebox{0.46\textwidth}{!}{ 
  \begin{tabular}{llccc}
    \toprule
    \textbf{Models} & \textbf{Setting}  & \textbf{AQuA} & \textbf{GSM8K} & \textbf{MATH} \\
    \midrule
    \multirow{4}{*}{Llama3-8B}
      & $k=2$ & $40.95$ &  $56.94$   & $13.20$  \\
      & $k=4$ & $42.91$ & $61.49$   & $15.28$  \\
      & $k=8$  & $46.46$ & $\textbf{62.62}$  & $\textbf{15.84}$  \\
      & $k=16$ & $\textbf{47.24}$ & $62.32$   & $15.64$  \\
    \midrule
    \multirow{4}{*}{Mistral-7B}
      & $k=2$  & $29.92$ & $40.10$   & $6.82$   \\
      & $k=4$  & $31.49$ & $47.46$   & $8.76$   \\
      & $k=8$  & $33.07$ & $\textbf{50.95}$  & $9.68$   \\
      & $k=16$ & $\textbf{33.46}$ & $50.19$   & $\textbf{9.82}$  \\
    \bottomrule
  \end{tabular}
}
\caption{Performance of \ours (State Reset) with varying values of $k$.}
\label{tab:different_sampling_times}
\end{table}

\section*{Limitations}
While our work introduces a new approach to mitigating the tail narrowing through \ours, there are still several limitations.
First, for computational efficiency, we do not scale the sampling in each iteration.
However, we conduct a series of experiments (\textsection~\ref{sec:performance_bottleneck}) to provide insights into how scaling the number of sampling can boost performance.
Second, following prior self-improvement works, we use binary signals for supervision based on final answer checks.
However, poor and spurious rationales while yielding correct answers may be utilized, which could hinder the improvement of reasoning ability. Filtering low-quality reasoning paths and ensuring the quality of self-generated data remains an area for further investigation.

\section*{Acknowledgment}
The authors wish to thank the anonymous reviewers for their helpful comments. This work was supported by the Science and Technology Commission of Shanghai Municipality (No. 24511103100), National Natural Science Foundation of China (No. 62476061,62206057,62076069), Shanghai Rising-Star Program (23QA1400200), Natural Science Foundation of Shanghai (23ZR1403500), Program of Shanghai Academic Research Leader under grant 22XD1401100. This research was also supported by Meituan.

\bibliography{ref}

\begin{thebibliography}{56}
\providecommand{\natexlab}[1]{#1}

\bibitem[{Adler et~al.(2024)Adler, Agarwal, Aithal, Anh, Bhattacharya, Brundyn, Casper, Catanzaro, Clay, Cohen, Das, Dattagupta, Delalleau, Derczynski, Dong, Egert, Evans, Ficek, Fridman, Ghosh, Ginsburg, Gitman, Grzegorzek, Hero, Huang, Jawa, Jennings, Jhunjhunwala, Kamalu, Khan, Kuchaiev, LeGresley, Li, Liu, Liu, Long, Mahabaleshwarkar, Majumdar, Maki, Martinez, de~Melo, Moshkov, Narayanan, Narenthiran, Navarro, Nguyen, Nitski, Noroozi, Nutheti, Parisien, Parmar, Patwary, Pawelec, Ping, Prabhumoye, Roy, Saar, Sabavat, Satheesh, Scowcroft, Sewall, Shamis, Shen, Shoeybi, Sizer, Smelyanskiy, Soares, Sreedhar, Su, Subramanian, Sun, Toshniwal, Wang, Wang, You, Zeng, Zhang, Zhang, Zhang, Zhang, and Zhu}]{nemotron}
Bo~Adler, Niket Agarwal, Ashwath Aithal, Dong~H. Anh, Pallab Bhattacharya, Annika Brundyn, Jared Casper, Bryan Catanzaro, Sharon Clay, Jonathan~M. Cohen, Sirshak Das, Ayush Dattagupta, Olivier Delalleau, Leon Derczynski, Yi~Dong, Daniel Egert, Ellie Evans, Aleksander Ficek, Denys Fridman, Shaona Ghosh, Boris Ginsburg, Igor Gitman, Tomasz Grzegorzek, Robert Hero, Jining Huang, Vibhu Jawa, Joseph Jennings, Aastha Jhunjhunwala, John Kamalu, Sadaf Khan, Oleksii Kuchaiev, Patrick LeGresley, Hui Li, Jiwei Liu, Zihan Liu, Eileen Long, Ameya~Sunil Mahabaleshwarkar, Somshubra Majumdar, James Maki, Miguel Martinez, Maer~Rodrigues de~Melo, Ivan Moshkov, Deepak Narayanan, Sean Narenthiran, Jesus Navarro, Phong Nguyen, Osvald Nitski, Vahid Noroozi, Guruprasad Nutheti, Christopher Parisien, Jupinder Parmar, Mostofa Patwary, Krzysztof Pawelec, Wei Ping, Shrimai Prabhumoye, Rajarshi Roy, Trisha Saar, Vasanth Rao~Naik Sabavat, Sanjeev Satheesh, Jane~Polak Scowcroft, Jason Sewall, Pavel Shamis, Gerald Shen, Mohammad Shoeybi,
  Dave Sizer, Misha Smelyanskiy, Felipe Soares, Makesh~Narsimhan Sreedhar, Dan Su, Sandeep Subramanian, Shengyang Sun, Shubham Toshniwal, Hao Wang, Zhilin Wang, Jiaxuan You, Jiaqi Zeng, Jimmy Zhang, Jing Zhang, Vivienne Zhang, Yian Zhang, and Chen Zhu. 2024.
\newblock \href {https://doi.org/10.48550/ARXIV.2406.11704} {Nemotron-4 340b technical report}.
\newblock \emph{CoRR}, abs/2406.11704.

\bibitem[{Alemohammad et~al.(2024)Alemohammad, Casco{-}Rodriguez, Luzi, Humayun, Babaei, LeJeune, Siahkoohi, and Baraniuk}]{self_consuming}
Sina Alemohammad, Josue Casco{-}Rodriguez, Lorenzo Luzi, Ahmed~Imtiaz Humayun, Hossein Babaei, Daniel LeJeune, Ali Siahkoohi, and Richard~G. Baraniuk. 2024.
\newblock \href {https://openreview.net/forum?id=ShjMHfmPs0} {Self-consuming generative models go {MAD}}.
\newblock In \emph{The Twelfth International Conference on Learning Representations, {ICLR} 2024, Vienna, Austria, May 7-11, 2024}. OpenReview.net.

\bibitem[{Amini et~al.(2019)Amini, Gabriel, Lin, Koncel{-}Kedziorski, Choi, and Hajishirzi}]{mathqa}
Aida Amini, Saadia Gabriel, Shanchuan Lin, Rik Koncel{-}Kedziorski, Yejin Choi, and Hannaneh Hajishirzi. 2019.
\newblock \href {https://doi.org/10.18653/V1/N19-1245} {Mathqa: Towards interpretable math word problem solving with operation-based formalisms}.
\newblock In \emph{Proceedings of the 2019 Conference of the North American Chapter of the Association for Computational Linguistics: Human Language Technologies, {NAACL-HLT} 2019, Minneapolis, MN, USA, June 2-7, 2019, Volume 1 (Long and Short Papers)}, pages 2357--2367. Association for Computational Linguistics.

\bibitem[{Bansal et~al.(2024)Bansal, Hosseini, Agarwal, Tran, and Kazemi}]{compute_sample}
Hritik Bansal, Arian Hosseini, Rishabh Agarwal, Vinh~Q. Tran, and Mehran Kazemi. 2024.
\newblock \href {https://doi.org/10.48550/ARXIV.2408.16737} {Smaller, weaker, yet better: Training {LLM} reasoners via compute-optimal sampling}.
\newblock \emph{CoRR}, abs/2408.16737.

\bibitem[{Chang(2023)}]{socratic}
Edward~Y. Chang. 2023.
\newblock \href {https://doi.org/10.1109/CCWC57344.2023.10099179} {Prompting large language models with the socratic method}.
\newblock In \emph{13th {IEEE} Annual Computing and Communication Workshop and Conference, {CCWC} 2023, Las Vegas, NV, USA, March 8-11, 2023}, pages 351--360. {IEEE}.

\bibitem[{Chang et~al.(2024)Chang, Zhan, Oertell, Brantley, Misra, Lee, and Sun}]{dataset_reset}
Jonathan~D. Chang, Wenhao Zhan, Owen Oertell, Kiant{\'{e}} Brantley, Dipendra Misra, Jason~D. Lee, and Wen Sun. 2024.
\newblock \href {https://doi.org/10.48550/ARXIV.2404.08495} {Dataset reset policy optimization for {RLHF}}.
\newblock \emph{CoRR}, abs/2404.08495.

\bibitem[{Chen et~al.(2023)Chen, Ma, Wang, and Cohen}]{pot}
Wenhu Chen, Xueguang Ma, Xinyi Wang, and William~W. Cohen. 2023.
\newblock \href {https://openreview.net/forum?id=YfZ4ZPt8zd} {Program of thoughts prompting: Disentangling computation from reasoning for numerical reasoning tasks}.
\newblock \emph{Trans. Mach. Learn. Res.}, 2023.

\bibitem[{Cobbe et~al.(2021)Cobbe, Kosaraju, Bavarian, Chen, Jun, Kaiser, Plappert, Tworek, Hilton, Nakano, Hesse, and Schulman}]{gsm8k}
Karl Cobbe, Vineet Kosaraju, Mohammad Bavarian, Mark Chen, Heewoo Jun, Lukasz Kaiser, Matthias Plappert, Jerry Tworek, Jacob Hilton, Reiichiro Nakano, Christopher Hesse, and John Schulman. 2021.
\newblock \href {https://arxiv.org/abs/2110.14168} {Training verifiers to solve math word problems}.
\newblock \emph{CoRR}, abs/2110.14168.

\bibitem[{Dohmatob et~al.(2024{\natexlab{a}})Dohmatob, Feng, Subramonian, and Kempe}]{strong_collapse}
Elvis Dohmatob, Yunzhen Feng, Arjun Subramonian, and Julia Kempe. 2024{\natexlab{a}}.
\newblock \href {https://arxiv.org/abs/2410.04840} {Strong model collapse}.
\newblock \emph{Preprint}, arXiv:2410.04840.

\bibitem[{Dohmatob et~al.(2024{\natexlab{b}})Dohmatob, Feng, Yang, Charton, and Kempe}]{tale_of_tails}
Elvis Dohmatob, Yunzhen Feng, Pu~Yang, Fran{\c{c}}ois Charton, and Julia Kempe. 2024{\natexlab{b}}.
\newblock \href {https://openreview.net/forum?id=KVvku47shW} {A tale of tails: Model collapse as a change of scaling laws}.
\newblock In \emph{Forty-first International Conference on Machine Learning, {ICML} 2024, Vienna, Austria, July 21-27, 2024}. OpenReview.net.

\bibitem[{Dong et~al.(2023{\natexlab{a}})Dong, Xiong, Goyal, Zhang, Chow, Pan, Diao, Zhang, Shum, and Zhang}]{raft}
Hanze Dong, Wei Xiong, Deepanshu Goyal, Yihan Zhang, Winnie Chow, Rui Pan, Shizhe Diao, Jipeng Zhang, Kashun Shum, and Tong Zhang. 2023{\natexlab{a}}.
\newblock \href {https://openreview.net/forum?id=m7p5O7zblY} {{RAFT:} reward ranked finetuning for generative foundation model alignment}.
\newblock \emph{Trans. Mach. Learn. Res.}, 2023.

\bibitem[{Dong et~al.(2023{\natexlab{b}})Dong, Dong, Xu, Zhou, Hao, Sui, and Wei}]{p_np}
Qingxiu Dong, Li~Dong, Ke~Xu, Guangyan Zhou, Yaru Hao, Zhifang Sui, and Furu Wei. 2023{\natexlab{b}}.
\newblock \href {https://doi.org/10.48550/ARXIV.2309.05689} {Large language model for science: {A} study on {P} vs. {NP}}.
\newblock \emph{CoRR}, abs/2309.05689.

\bibitem[{Dubey et~al.(2024)Dubey, Jauhri, Pandey, Kadian, Al{-}Dahle, Letman, Mathur, Schelten, Yang, Fan, Goyal, Hartshorn, Yang, Mitra, Sravankumar, Korenev, Hinsvark, Rao, Zhang, Rodriguez, Gregerson, Spataru, Rozi{\`{e}}re, Biron, Tang, Chern, Caucheteux, Nayak, Bi, Marra, McConnell, Keller, Touret, Wu, Wong, Ferrer, Nikolaidis, Allonsius, Song, Pintz, Livshits, Esiobu, Choudhary, Mahajan, Garcia{-}Olano, Perino, Hupkes, Lakomkin, AlBadawy, Lobanova, Dinan, Smith, Radenovic, Zhang, Synnaeve, Lee, Anderson, Nail, Mialon, Pang, Cucurell, Nguyen, Korevaar, Xu, Touvron, Zarov, Ibarra, Kloumann, Misra, Evtimov, Copet, Lee, Geffert, Vranes, Park, Mahadeokar, Shah, van~der Linde, Billock, Hong, Lee, Fu, Chi, Huang, Liu, Wang, Yu, Bitton, Spisak, Park, Rocca, Johnstun, Saxe, Jia, Alwala, Upasani, Plawiak, Li, Heafield, Stone, and et~al.}]{llama3}
Abhimanyu Dubey, Abhinav Jauhri, Abhinav Pandey, Abhishek Kadian, Ahmad Al{-}Dahle, Aiesha Letman, Akhil Mathur, Alan Schelten, Amy Yang, Angela Fan, Anirudh Goyal, Anthony Hartshorn, Aobo Yang, Archi Mitra, Archie Sravankumar, Artem Korenev, Arthur Hinsvark, Arun Rao, Aston Zhang, Aur{\'{e}}lien Rodriguez, Austen Gregerson, Ava Spataru, Baptiste Rozi{\`{e}}re, Bethany Biron, Binh Tang, Bobbie Chern, Charlotte Caucheteux, Chaya Nayak, Chloe Bi, Chris Marra, Chris McConnell, Christian Keller, Christophe Touret, Chunyang Wu, Corinne Wong, Cristian~Canton Ferrer, Cyrus Nikolaidis, Damien Allonsius, Daniel Song, Danielle Pintz, Danny Livshits, David Esiobu, Dhruv Choudhary, Dhruv Mahajan, Diego Garcia{-}Olano, Diego Perino, Dieuwke Hupkes, Egor Lakomkin, Ehab AlBadawy, Elina Lobanova, Emily Dinan, Eric~Michael Smith, Filip Radenovic, Frank Zhang, Gabriel Synnaeve, Gabrielle Lee, Georgia~Lewis Anderson, Graeme Nail, Gr{\'{e}}goire Mialon, Guan Pang, Guillem Cucurell, Hailey Nguyen, Hannah Korevaar, Hu~Xu, Hugo
  Touvron, Iliyan Zarov, Imanol~Arrieta Ibarra, Isabel~M. Kloumann, Ishan Misra, Ivan Evtimov, Jade Copet, Jaewon Lee, Jan Geffert, Jana Vranes, Jason Park, Jay Mahadeokar, Jeet Shah, Jelmer van~der Linde, Jennifer Billock, Jenny Hong, Jenya Lee, Jeremy Fu, Jianfeng Chi, Jianyu Huang, Jiawen Liu, Jie Wang, Jiecao Yu, Joanna Bitton, Joe Spisak, Jongsoo Park, Joseph Rocca, Joshua Johnstun, Joshua Saxe, Junteng Jia, Kalyan~Vasuden Alwala, Kartikeya Upasani, Kate Plawiak, Ke~Li, Kenneth Heafield, Kevin Stone, and et~al. 2024.
\newblock \href {https://doi.org/10.48550/ARXIV.2407.21783} {The llama 3 herd of models}.
\newblock \emph{CoRR}, abs/2407.21783.

\bibitem[{Fu et~al.(2023)Fu, Peng, Sabharwal, Clark, and Khot}]{complexity_cot}
Yao Fu, Hao Peng, Ashish Sabharwal, Peter Clark, and Tushar Khot. 2023.
\newblock \href {https://openreview.net/forum?id=yf1icZHC-l9} {Complexity-based prompting for multi-step reasoning}.
\newblock In \emph{The Eleventh International Conference on Learning Representations, {ICLR} 2023, Kigali, Rwanda, May 1-5, 2023}. OpenReview.net.

\bibitem[{G{\"{u}}l{\c{c}}ehre et~al.(2023)G{\"{u}}l{\c{c}}ehre, Paine, Srinivasan, Konyushkova, Weerts, Sharma, Siddhant, Ahern, Wang, Gu, Macherey, Doucet, Firat, and de~Freitas}]{rest}
{\c{C}}aglar G{\"{u}}l{\c{c}}ehre, Tom~Le Paine, Srivatsan Srinivasan, Ksenia Konyushkova, Lotte Weerts, Abhishek Sharma, Aditya Siddhant, Alex Ahern, Miaosen Wang, Chenjie Gu, Wolfgang Macherey, Arnaud Doucet, Orhan Firat, and Nando de~Freitas. 2023.
\newblock \href {https://doi.org/10.48550/ARXIV.2308.08998} {Reinforced self-training (rest) for language modeling}.
\newblock \emph{CoRR}, abs/2308.08998.

\bibitem[{Guo et~al.(2024{\natexlab{a}})Guo, Zhu, Yang, Xie, Dong, Zhang, Chen, Bi, Wu, Li, Luo, Xiong, and Liang}]{deepseek_coder}
Daya Guo, Qihao Zhu, Dejian Yang, Zhenda Xie, Kai Dong, Wentao Zhang, Guanting Chen, Xiao Bi, Y.~Wu, Y.~K. Li, Fuli Luo, Yingfei Xiong, and Wenfeng Liang. 2024{\natexlab{a}}.
\newblock \href {https://doi.org/10.48550/ARXIV.2401.14196} {Deepseek-coder: When the large language model meets programming - the rise of code intelligence}.
\newblock \emph{CoRR}, abs/2401.14196.

\bibitem[{Guo et~al.(2024{\natexlab{b}})Guo, Shang, Vazirgiannis, and Clavel}]{Linguistic_Diversity}
Yanzhu Guo, Guokan Shang, Michalis Vazirgiannis, and Chlo{\'{e}} Clavel. 2024{\natexlab{b}}.
\newblock \href {https://doi.org/10.18653/V1/2024.FINDINGS-NAACL.228} {The curious decline of linguistic diversity: Training language models on synthetic text}.
\newblock In \emph{Findings of the Association for Computational Linguistics: {NAACL} 2024, Mexico City, Mexico, June 16-21, 2024}, pages 3589--3604. Association for Computational Linguistics.

\bibitem[{Haluptzok et~al.(2023)Haluptzok, Bowers, and Kalai}]{program_better}
Patrick Haluptzok, Matthew Bowers, and Adam~Tauman Kalai. 2023.
\newblock \href {https://openreview.net/forum?id=SaRj2ka1XZ3} {Language models can teach themselves to program better}.
\newblock In \emph{The Eleventh International Conference on Learning Representations, {ICLR} 2023, Kigali, Rwanda, May 1-5, 2023}. OpenReview.net.

\bibitem[{Hendrycks et~al.(2021)Hendrycks, Burns, Kadavath, Arora, Basart, Tang, Song, and Steinhardt}]{MATH}
Dan Hendrycks, Collin Burns, Saurav Kadavath, Akul Arora, Steven Basart, Eric Tang, Dawn Song, and Jacob Steinhardt. 2021.
\newblock \href {https://datasets-benchmarks-proceedings.neurips.cc/paper/2021/hash/be83ab3ecd0db773eb2dc1b0a17836a1-Abstract-round2.html} {Measuring mathematical problem solving with the {MATH} dataset}.
\newblock In \emph{Proceedings of the Neural Information Processing Systems Track on Datasets and Benchmarks 1, NeurIPS Datasets and Benchmarks 2021, December 2021, virtual}.

\bibitem[{Hosseini et~al.(2024)Hosseini, Yuan, Malkin, Courville, Sordoni, and Agarwal}]{vstar}
Arian Hosseini, Xingdi Yuan, Nikolay Malkin, Aaron~C. Courville, Alessandro Sordoni, and Rishabh Agarwal. 2024.
\newblock \href {https://doi.org/10.48550/ARXIV.2402.06457} {V-star: Training verifiers for self-taught reasoners}.
\newblock \emph{CoRR}, abs/2402.06457.

\bibitem[{Huang et~al.(2023)Huang, Gu, Hou, Wu, Wang, Yu, and Han}]{llm_self_improve}
Jiaxin Huang, Shixiang Gu, Le~Hou, Yuexin Wu, Xuezhi Wang, Hongkun Yu, and Jiawei Han. 2023.
\newblock \href {https://doi.org/10.18653/V1/2023.EMNLP-MAIN.67} {Large language models can self-improve}.
\newblock In \emph{Proceedings of the 2023 Conference on Empirical Methods in Natural Language Processing, {EMNLP} 2023, Singapore, December 6-10, 2023}, pages 1051--1068. Association for Computational Linguistics.

\bibitem[{Jiang et~al.(2023)Jiang, Sablayrolles, Mensch, Bamford, Chaplot, de~Las~Casas, Bressand, Lengyel, Lample, Saulnier, Lavaud, Lachaux, Stock, Scao, Lavril, Wang, Lacroix, and Sayed}]{mistral}
Albert~Q. Jiang, Alexandre Sablayrolles, Arthur Mensch, Chris Bamford, Devendra~Singh Chaplot, Diego de~Las~Casas, Florian Bressand, Gianna Lengyel, Guillaume Lample, Lucile Saulnier, L{\'{e}}lio~Renard Lavaud, Marie{-}Anne Lachaux, Pierre Stock, Teven~Le Scao, Thibaut Lavril, Thomas Wang, Timoth{\'{e}}e Lacroix, and William~El Sayed. 2023.
\newblock \href {https://doi.org/10.48550/ARXIV.2310.06825} {Mistral 7b}.
\newblock \emph{CoRR}, abs/2310.06825.

\bibitem[{Kaplan et~al.(2020)Kaplan, McCandlish, Henighan, Brown, Chess, Child, Gray, Radford, Wu, and Amodei}]{scaling_law}
Jared Kaplan, Sam McCandlish, Tom Henighan, Tom~B. Brown, Benjamin Chess, Rewon Child, Scott Gray, Alec Radford, Jeffrey Wu, and Dario Amodei. 2020.
\newblock \href {https://arxiv.org/abs/2001.08361} {Scaling laws for neural language models}.
\newblock \emph{CoRR}, abs/2001.08361.

\bibitem[{Kojima et~al.(2022)Kojima, Gu, Reid, Matsuo, and Iwasawa}]{zerocot}
Takeshi Kojima, Shixiang~Shane Gu, Machel Reid, Yutaka Matsuo, and Yusuke Iwasawa. 2022.
\newblock \href {http://papers.nips.cc/paper\_files/paper/2022/hash/8bb0d291acd4acf06ef112099c16f326-Abstract-Conference.html} {Large language models are zero-shot reasoners}.
\newblock In \emph{Advances in Neural Information Processing Systems 35: Annual Conference on Neural Information Processing Systems 2022, NeurIPS 2022, New Orleans, LA, USA, November 28 - December 9, 2022}.

\bibitem[{Kwon et~al.(2023)Kwon, Li, Zhuang, Sheng, Zheng, Yu, Gonzalez, Zhang, and Stoica}]{vllm}
Woosuk Kwon, Zhuohan Li, Siyuan Zhuang, Ying Sheng, Lianmin Zheng, Cody~Hao Yu, Joseph Gonzalez, Hao Zhang, and Ion Stoica. 2023.
\newblock \href {https://doi.org/10.1145/3600006.3613165} {Efficient memory management for large language model serving with pagedattention}.
\newblock In \emph{Proceedings of the 29th Symposium on Operating Systems Principles, {SOSP} 2023, Koblenz, Germany, October 23-26, 2023}, pages 611--626. {ACM}.

\bibitem[{Lanham et~al.(2023)Lanham, Chen, Radhakrishnan, Steiner, Denison, Hernandez, Li, Durmus, Hubinger, Kernion, Lukosiute, Nguyen, Cheng, Joseph, Schiefer, Rausch, Larson, McCandlish, Kundu, Kadavath, Yang, Henighan, Maxwell, Telleen{-}Lawton, Hume, Hatfield{-}Dodds, Kaplan, Brauner, Bowman, and Perez}]{cot_faithfulness}
Tamera Lanham, Anna Chen, Ansh Radhakrishnan, Benoit Steiner, Carson Denison, Danny Hernandez, Dustin Li, Esin Durmus, Evan Hubinger, Jackson Kernion, Kamile Lukosiute, Karina Nguyen, Newton Cheng, Nicholas Joseph, Nicholas Schiefer, Oliver Rausch, Robin Larson, Sam McCandlish, Sandipan Kundu, Saurav Kadavath, Shannon Yang, Thomas Henighan, Timothy Maxwell, Timothy Telleen{-}Lawton, Tristan Hume, Zac Hatfield{-}Dodds, Jared Kaplan, Jan Brauner, Samuel~R. Bowman, and Ethan Perez. 2023.
\newblock \href {https://doi.org/10.48550/ARXIV.2307.13702} {Measuring faithfulness in chain-of-thought reasoning}.
\newblock \emph{CoRR}, abs/2307.13702.

\bibitem[{Lightman et~al.(2024)Lightman, Kosaraju, Burda, Edwards, Baker, Lee, Leike, Schulman, Sutskever, and Cobbe}]{verify}
Hunter Lightman, Vineet Kosaraju, Yuri Burda, Harrison Edwards, Bowen Baker, Teddy Lee, Jan Leike, John Schulman, Ilya Sutskever, and Karl Cobbe. 2024.
\newblock \href {https://openreview.net/forum?id=v8L0pN6EOi} {Let's verify step by step}.
\newblock In \emph{The Twelfth International Conference on Learning Representations, {ICLR} 2024, Vienna, Austria, May 7-11, 2024}. OpenReview.net.

\bibitem[{Ling et~al.(2017)Ling, Yogatama, Dyer, and Blunsom}]{AQuA}
Wang Ling, Dani Yogatama, Chris Dyer, and Phil Blunsom. 2017.
\newblock \href {https://doi.org/10.18653/V1/P17-1015} {Program induction by rationale generation: Learning to solve and explain algebraic word problems}.
\newblock In \emph{Proceedings of the 55th Annual Meeting of the Association for Computational Linguistics, {ACL} 2017, Vancouver, Canada, July 30 - August 4, Volume 1: Long Papers}, pages 158--167. Association for Computational Linguistics.

\bibitem[{Liu et~al.(2024)Liu, Zeng, He, Jiang, and He}]{deita}
Wei Liu, Weihao Zeng, Keqing He, Yong Jiang, and Junxian He. 2024.
\newblock \href {https://openreview.net/forum?id=BTKAeLqLMw} {What makes good data for alignment? {A} comprehensive study of automatic data selection in instruction tuning}.
\newblock In \emph{The Twelfth International Conference on Learning Representations, {ICLR} 2024, Vienna, Austria, May 7-11, 2024}. OpenReview.net.

\bibitem[{Patel et~al.(2021)Patel, Bhattamishra, and Goyal}]{svamp}
Arkil Patel, Satwik Bhattamishra, and Navin Goyal. 2021.
\newblock \href {https://doi.org/10.18653/V1/2021.NAACL-MAIN.168} {Are {NLP} models really able to solve simple math word problems?}
\newblock In \emph{Proceedings of the 2021 Conference of the North American Chapter of the Association for Computational Linguistics: Human Language Technologies, {NAACL-HLT} 2021, Online, June 6-11, 2021}, pages 2080--2094. Association for Computational Linguistics.

\bibitem[{Qi et~al.(2024)Qi, Ma, Xu, Zhang, Yang, and Yang}]{rstar}
Zhenting Qi, Mingyuan Ma, Jiahang Xu, Li~Lyna Zhang, Fan Yang, and Mao Yang. 2024.
\newblock \href {https://doi.org/10.48550/ARXIV.2408.06195} {Mutual reasoning makes smaller llms stronger problem-solvers}.
\newblock \emph{CoRR}, abs/2408.06195.

\bibitem[{Rozi{\`{e}}re et~al.(2023)Rozi{\`{e}}re, Gehring, Gloeckle, Sootla, Gat, Tan, Adi, Liu, Remez, Rapin, Kozhevnikov, Evtimov, Bitton, Bhatt, Canton{-}Ferrer, Grattafiori, Xiong, D{\'{e}}fossez, Copet, Azhar, Touvron, Martin, Usunier, Scialom, and Synnaeve}]{codellama}
Baptiste Rozi{\`{e}}re, Jonas Gehring, Fabian Gloeckle, Sten Sootla, Itai Gat, Xiaoqing~Ellen Tan, Yossi Adi, Jingyu Liu, Tal Remez, J{\'{e}}r{\'{e}}my Rapin, Artyom Kozhevnikov, Ivan Evtimov, Joanna Bitton, Manish Bhatt, Cristian Canton{-}Ferrer, Aaron Grattafiori, Wenhan Xiong, Alexandre D{\'{e}}fossez, Jade Copet, Faisal Azhar, Hugo Touvron, Louis Martin, Nicolas Usunier, Thomas Scialom, and Gabriel Synnaeve. 2023.
\newblock \href {https://doi.org/10.48550/ARXIV.2308.12950} {Code llama: Open foundation models for code}.
\newblock \emph{CoRR}, abs/2308.12950.

\bibitem[{Schaal(1996)}]{learning_from_demos}
Stefan Schaal. 1996.
\newblock \href {http://papers.nips.cc/paper/1224-learning-from-demonstration} {Learning from demonstration}.
\newblock In \emph{Advances in Neural Information Processing Systems 9, NIPS, Denver, CO, USA, December 2-5, 1996}, pages 1040--1046. {MIT} Press.

\bibitem[{Setlur et~al.(2024)Setlur, Garg, Geng, Garg, Smith, and Kumar}]{rl_on_incorrect}
Amrith Setlur, Saurabh Garg, Xinyang Geng, Naman Garg, Virginia Smith, and Aviral Kumar. 2024.
\newblock \href {https://doi.org/10.48550/ARXIV.2406.14532} {{RL} on incorrect synthetic data scales the efficiency of {LLM} math reasoning by eight-fold}.
\newblock \emph{CoRR}, abs/2406.14532.

\bibitem[{Shao et~al.(2024)Shao, Wang, Zhu, Xu, Song, Zhang, Li, Wu, and Guo}]{deepseek}
Zhihong Shao, Peiyi Wang, Qihao Zhu, Runxin Xu, Junxiao Song, Mingchuan Zhang, Y.~K. Li, Y.~Wu, and Daya Guo. 2024.
\newblock \href {https://doi.org/10.48550/ARXIV.2402.03300} {Deepseekmath: Pushing the limits of mathematical reasoning in open language models}.
\newblock \emph{CoRR}, abs/2402.03300.

\bibitem[{Shumailov et~al.(2023)Shumailov, Shumaylov, Zhao, Gal, Papernot, and Anderson}]{curse_recursion}
Ilia Shumailov, Zakhar Shumaylov, Yiren Zhao, Yarin Gal, Nicolas Papernot, and Ross~J. Anderson. 2023.
\newblock \href {https://doi.org/10.48550/ARXIV.2305.17493} {The curse of recursion: Training on generated data makes models forget}.
\newblock \emph{CoRR}, abs/2305.17493.

\bibitem[{Singh et~al.(2024)Singh, Co{-}Reyes, Agarwal, Anand, Patil, Garcia, Liu, Harrison, Lee, Xu, Parisi, Kumar, Alemi, Rizkowsky, Nova, Adlam, Bohnet, Elsayed, Sedghi, Mordatch, Simpson, Gur, Snoek, Pennington, Hron, Kenealy, Swersky, Mahajan, Culp, Xiao, Bileschi, Constant, Novak, Liu, Warkentin, Qian, Bansal, Dyer, Neyshabur, Sohl{-}Dickstein, and Fiedel}]{rest_em}
Avi Singh, John~D. Co{-}Reyes, Rishabh Agarwal, Ankesh Anand, Piyush Patil, Xavier Garcia, Peter~J. Liu, James Harrison, Jaehoon Lee, Kelvin Xu, Aaron~T. Parisi, Abhishek Kumar, Alexander~A. Alemi, Alex Rizkowsky, Azade Nova, Ben Adlam, Bernd Bohnet, Gamaleldin~Fathy Elsayed, Hanie Sedghi, Igor Mordatch, Isabelle Simpson, Izzeddin Gur, Jasper Snoek, Jeffrey Pennington, Jiri Hron, Kathleen Kenealy, Kevin Swersky, Kshiteej Mahajan, Laura Culp, Lechao Xiao, Maxwell~L. Bileschi, Noah Constant, Roman Novak, Rosanne Liu, Tris Warkentin, Yundi Qian, Yamini Bansal, Ethan Dyer, Behnam Neyshabur, Jascha Sohl{-}Dickstein, and Noah Fiedel. 2024.
\newblock \href {https://openreview.net/forum?id=lNAyUngGFK} {Beyond human data: Scaling self-training for problem-solving with language models}.
\newblock \emph{Trans. Mach. Learn. Res.}, 2024.

\bibitem[{Sorscher et~al.(2022)Sorscher, Geirhos, Shekhar, Ganguli, and Morcos}]{beyond_neural_scaling}
Ben Sorscher, Robert Geirhos, Shashank Shekhar, Surya Ganguli, and Ari Morcos. 2022.
\newblock \href {http://papers.nips.cc/paper\_files/paper/2022/hash/7b75da9b61eda40fa35453ee5d077df6-Abstract-Conference.html} {Beyond neural scaling laws: beating power law scaling via data pruning}.
\newblock In \emph{Advances in Neural Information Processing Systems 35: Annual Conference on Neural Information Processing Systems 2022, NeurIPS 2022, New Orleans, LA, USA, November 28 - December 9, 2022}.

\bibitem[{Suay and Chernova(2011)}]{suay2011effect}
Halit~Bener Suay and Sonia Chernova. 2011.
\newblock Effect of human guidance and state space size on interactive reinforcement learning.
\newblock In \emph{2011 Ro-Man}, pages 1--6. IEEE.

\bibitem[{Subramanian et~al.(2016)Subramanian, Jr., and Thomaz}]{exploration_from_demos}
Kaushik Subramanian, Charles Lee~Isbell Jr., and Andrea~Lockerd Thomaz. 2016.
\newblock \href {http://dl.acm.org/citation.cfm?id=2936990} {Exploration from demonstration for interactive reinforcement learning}.
\newblock In \emph{Proceedings of the 2016 International Conference on Autonomous Agents {\&} Multiagent Systems, Singapore, May 9-13, 2016}, pages 447--456. {ACM}.

\bibitem[{Tong et~al.(2024)Tong, Zhang, Wang, Wu, and He}]{dart_math}
Yuxuan Tong, Xiwen Zhang, Rui Wang, Ruidong Wu, and Junxian He. 2024.
\newblock \href {https://doi.org/10.48550/ARXIV.2407.13690} {Dart-math: Difficulty-aware rejection tuning for mathematical problem-solving}.
\newblock \emph{CoRR}, abs/2407.13690.

\bibitem[{Touvron et~al.(2023)Touvron, Martin, Stone, Albert, Almahairi, Babaei, Bashlykov, Batra, Bhargava, Bhosale, Bikel, Blecher, Canton{-}Ferrer, Chen, Cucurull, Esiobu, Fernandes, Fu, Fu, Fuller, Gao, Goswami, Goyal, Hartshorn, Hosseini, Hou, Inan, Kardas, Kerkez, Khabsa, Kloumann, Korenev, Koura, Lachaux, Lavril, Lee, Liskovich, Lu, Mao, Martinet, Mihaylov, Mishra, Molybog, Nie, Poulton, Reizenstein, Rungta, Saladi, Schelten, Silva, Smith, Subramanian, Tan, Tang, Taylor, Williams, Kuan, Xu, Yan, Zarov, Zhang, Fan, Kambadur, Narang, Rodriguez, Stojnic, Edunov, and Scialom}]{llama2}
Hugo Touvron, Louis Martin, Kevin Stone, Peter Albert, Amjad Almahairi, Yasmine Babaei, Nikolay Bashlykov, Soumya Batra, Prajjwal Bhargava, Shruti Bhosale, Dan Bikel, Lukas Blecher, Cristian Canton{-}Ferrer, Moya Chen, Guillem Cucurull, David Esiobu, Jude Fernandes, Jeremy Fu, Wenyin Fu, Brian Fuller, Cynthia Gao, Vedanuj Goswami, Naman Goyal, Anthony Hartshorn, Saghar Hosseini, Rui Hou, Hakan Inan, Marcin Kardas, Viktor Kerkez, Madian Khabsa, Isabel Kloumann, Artem Korenev, Punit~Singh Koura, Marie{-}Anne Lachaux, Thibaut Lavril, Jenya Lee, Diana Liskovich, Yinghai Lu, Yuning Mao, Xavier Martinet, Todor Mihaylov, Pushkar Mishra, Igor Molybog, Yixin Nie, Andrew Poulton, Jeremy Reizenstein, Rashi Rungta, Kalyan Saladi, Alan Schelten, Ruan Silva, Eric~Michael Smith, Ranjan Subramanian, Xiaoqing~Ellen Tan, Binh Tang, Ross Taylor, Adina Williams, Jian~Xiang Kuan, Puxin Xu, Zheng Yan, Iliyan Zarov, Yuchen Zhang, Angela Fan, Melanie Kambadur, Sharan Narang, Aur{\'{e}}lien Rodriguez, Robert Stojnic, Sergey Edunov,
  and Thomas Scialom. 2023.
\newblock \href {https://doi.org/10.48550/ARXIV.2307.09288} {Llama 2: Open foundation and fine-tuned chat models}.
\newblock \emph{CoRR}, abs/2307.09288.

\bibitem[{Wei et~al.(2022{\natexlab{a}})Wei, Tay, Bommasani, Raffel, Zoph, Borgeaud, Yogatama, Bosma, Zhou, Metzler, Chi, Hashimoto, Vinyals, Liang, Dean, and Fedus}]{wei2022emergent}
Jason Wei, Yi~Tay, Rishi Bommasani, Colin Raffel, Barret Zoph, Sebastian Borgeaud, Dani Yogatama, Maarten Bosma, Denny Zhou, Donald Metzler, Ed~H. Chi, Tatsunori Hashimoto, Oriol Vinyals, Percy Liang, Jeff Dean, and William Fedus. 2022{\natexlab{a}}.
\newblock \href {https://openreview.net/forum?id=yzkSU5zdwD} {Emergent abilities of large language models}.
\newblock \emph{Trans. Mach. Learn. Res.}, 2022.

\bibitem[{Wei et~al.(2022{\natexlab{b}})Wei, Wang, Schuurmans, Bosma, Xia, Chi, Le, Zhou et~al.}]{wei2022chain}
Jason Wei, Xuezhi Wang, Dale Schuurmans, Maarten Bosma, Fei Xia, Ed~Chi, Quoc~V Le, Denny Zhou, et~al. 2022{\natexlab{b}}.
\newblock Chain-of-thought prompting elicits reasoning in large language models.
\newblock \emph{Advances in neural information processing systems}, 35:24824--24837.

\bibitem[{Wu et~al.(2024)Wu, Li, and Liu}]{progress_or_regress}
Ting Wu, Xuefeng Li, and Pengfei Liu. 2024.
\newblock \href {https://doi.org/10.48550/ARXIV.2407.05013} {Progress or regress? self-improvement reversal in post-training}.
\newblock \emph{CoRR}, abs/2407.05013.

\bibitem[{Xi et~al.(2024)Xi, Chen, Hong, Jin, Zheng, He, Ding, Liu, Guo, Wang, Guo, Shen, Fan, Zhou, Dou, Wang, Zhang, Sun, Gui, Zhang, and Huang}]{r3}
Zhiheng Xi, Wenxiang Chen, Boyang Hong, Senjie Jin, Rui Zheng, Wei He, Yiwen Ding, Shichun Liu, Xin Guo, Junzhe Wang, Honglin Guo, Wei Shen, Xiaoran Fan, Yuhao Zhou, Shihan Dou, Xiao Wang, Xinbo Zhang, Peng Sun, Tao Gui, Qi~Zhang, and Xuanjing Huang. 2024.
\newblock \href {https://openreview.net/forum?id=t82Y3fmRtk} {Training large language models for reasoning through reverse curriculum reinforcement learning}.
\newblock In \emph{Forty-first International Conference on Machine Learning, {ICML} 2024, Vienna, Austria, July 21-27, 2024}. OpenReview.net.

\bibitem[{Xie et~al.(2023)Xie, Kawaguchi, Zhao, Zhao, Kan, He, and Xie}]{xie23self}
Yuxi Xie, Kenji Kawaguchi, Yiran Zhao, James~Xu Zhao, Min{-}Yen Kan, Junxian He, and Michael~Qizhe Xie. 2023.
\newblock \href {http://papers.nips.cc/paper\_files/paper/2023/hash/81fde95c4dc79188a69ce5b24d63010b-Abstract-Conference.html} {Self-evaluation guided beam search for reasoning}.
\newblock In \emph{Advances in Neural Information Processing Systems 36: Annual Conference on Neural Information Processing Systems 2023, NeurIPS 2023, New Orleans, LA, USA, December 10 - 16, 2023}.

\bibitem[{Yang et~al.(2024)Yang, Pang, Feng, Wang, Chen, Zhu, and Liu}]{sdft}
Zhaorui Yang, Tianyu Pang, Haozhe Feng, Han Wang, Wei Chen, Minfeng Zhu, and Qian Liu. 2024.
\newblock \href {https://doi.org/10.18653/V1/2024.ACL-LONG.58} {Self-distillation bridges distribution gap in language model fine-tuning}.
\newblock In \emph{Proceedings of the 62nd Annual Meeting of the Association for Computational Linguistics (Volume 1: Long Papers), {ACL} 2024, Bangkok, Thailand, August 11-16, 2024}, pages 1028--1043. Association for Computational Linguistics.

\bibitem[{Yu et~al.(2024)Yu, Jiang, Shi, Yu, Liu, Zhang, Kwok, Li, Weller, and Liu}]{metamath}
Longhui Yu, Weisen Jiang, Han Shi, Jincheng Yu, Zhengying Liu, Yu~Zhang, James~T. Kwok, Zhenguo Li, Adrian Weller, and Weiyang Liu. 2024.
\newblock \href {https://openreview.net/forum?id=N8N0hgNDRt} {Metamath: Bootstrap your own mathematical questions for large language models}.
\newblock In \emph{The Twelfth International Conference on Learning Representations, {ICLR} 2024, Vienna, Austria, May 7-11, 2024}. OpenReview.net.

\bibitem[{Yu et~al.(2023)Yu, Zhuang, Zhang, Meng, Ratner, Krishna, Shen, and Zhang}]{tale_of_diversity_and_bias}
Yue Yu, Yuchen Zhuang, Jieyu Zhang, Yu~Meng, Alexander~J. Ratner, Ranjay Krishna, Jiaming Shen, and Chao Zhang. 2023.
\newblock \href {http://papers.nips.cc/paper\_files/paper/2023/hash/ae9500c4f5607caf2eff033c67daa9d7-Abstract-Datasets\_and\_Benchmarks.html} {Large language model as attributed training data generator: {A} tale of diversity and bias}.
\newblock In \emph{Advances in Neural Information Processing Systems 36: Annual Conference on Neural Information Processing Systems 2023, NeurIPS 2023, New Orleans, LA, USA, December 10 - 16, 2023}.

\bibitem[{Yuan et~al.(2024)Yuan, Pang, Cho, Li, Sukhbaatar, Xu, and Weston}]{self_rewarding}
Weizhe Yuan, Richard~Yuanzhe Pang, Kyunghyun Cho, Xian Li, Sainbayar Sukhbaatar, Jing Xu, and Jason Weston. 2024.
\newblock \href {https://openreview.net/forum?id=0NphYCmgua} {Self-rewarding language models}.
\newblock In \emph{Forty-first International Conference on Machine Learning, {ICML} 2024, Vienna, Austria, July 21-27, 2024}. OpenReview.net.

\bibitem[{Yuan et~al.(2023)Yuan, Yuan, Li, Dong, Tan, and Zhou}]{rft}
Zheng Yuan, Hongyi Yuan, Chengpeng Li, Guanting Dong, Chuanqi Tan, and Chang Zhou. 2023.
\newblock \href {https://doi.org/10.48550/ARXIV.2308.01825} {Scaling relationship on learning mathematical reasoning with large language models}.
\newblock \emph{CoRR}, abs/2308.01825.

\bibitem[{Yue et~al.(2024)Yue, Qu, Zhang, Fu, Huang, Sun, Su, and Chen}]{mammoth}
Xiang Yue, Xingwei Qu, Ge~Zhang, Yao Fu, Wenhao Huang, Huan Sun, Yu~Su, and Wenhu Chen. 2024.
\newblock \href {https://openreview.net/forum?id=yLClGs770I} {Mammoth: Building math generalist models through hybrid instruction tuning}.
\newblock In \emph{The Twelfth International Conference on Learning Representations, {ICLR} 2024, Vienna, Austria, May 7-11, 2024}. OpenReview.net.

\bibitem[{Zelikman et~al.(2022)Zelikman, Wu, Mu, and Goodman}]{star}
Eric Zelikman, Yuhuai Wu, Jesse Mu, and Noah Goodman. 2022.
\newblock Star: Bootstrapping reasoning with reasoning.
\newblock \emph{Advances in Neural Information Processing Systems}, 35:15476--15488.

\bibitem[{Zhang et~al.(2023)Zhang, Li, Cui, Cai, Liu, Fu, Huang, Zhao, Zhang, Chen, Wang, Luu, Bi, Shi, and Shi}]{zhang23hallucination}
Yue Zhang, Yafu Li, Leyang Cui, Deng Cai, Lemao Liu, Tingchen Fu, Xinting Huang, Enbo Zhao, Yu~Zhang, Yulong Chen, Longyue Wang, Anh~Tuan Luu, Wei Bi, Freda Shi, and Shuming Shi. 2023.
\newblock \href {https://doi.org/10.48550/ARXIV.2309.01219} {Siren's song in the {AI} ocean: {A} survey on hallucination in large language models}.
\newblock \emph{CoRR}, abs/2309.01219.

\bibitem[{Zhao et~al.(2023)Zhao, Zhou, Li, Tang, Wang, Hou, Min, Zhang, Zhang, Dong, Du, Yang, Chen, Chen, Jiang, Ren, Li, Tang, Liu, Liu, Nie, and Wen}]{llm_survey}
Wayne~Xin Zhao, Kun Zhou, Junyi Li, Tianyi Tang, Xiaolei Wang, Yupeng Hou, Yingqian Min, Beichen Zhang, Junjie Zhang, Zican Dong, Yifan Du, Chen Yang, Yushuo Chen, Zhipeng Chen, Jinhao Jiang, Ruiyang Ren, Yifan Li, Xinyu Tang, Zikang Liu, Peiyu Liu, Jian{-}Yun Nie, and Ji{-}Rong Wen. 2023.
\newblock \href {https://doi.org/10.48550/ARXIV.2303.18223} {A survey of large language models}.
\newblock \emph{CoRR}, abs/2303.18223.

\end{thebibliography}

\section*{Appendix}
\renewcommand{\thesubsection}{\Alph{subsection}}

\subsection{Dataset Details}
\label{app:data}
\begin{table}[ht]
\centering
\setlength{\tabcolsep}{3pt}
\resizebox{0.48\textwidth}{!}{%
\begin{tabular}{lccccccc}
\toprule
            & AQuA & GSM8K & MATH & MathQA & SVAMP & Thm.QA \\
\hline
\# CoT Train & $3000$ & $3000$  & $4000$ & -      & -     & -         \\
\# PoT Train & $1961$ & $3000$  & $4000$ & -      & -     & -         \\
\# Test      & $254$  & $1319$  & $5000$ & $2985$   & $1000$  & $800$       \\
\bottomrule
\end{tabular}
}
\caption{Dataset statistics of the train and test set.}
\label{tab:data}
\end{table}
For the training set, we randomly select a subset from the extensive datasets provided by MathInstruct \cite{mammoth}, including AQuA, GSM8K, and MATH. The test set is consistent with MathInstruct. The specific data quantities are shown in Table \ref{tab:data}. We comply with the license for the use of these datasets in our work.

\subsection{Error Patterns} \label{app:error_patterns}

Here, we analyze two classic types of errors in model-generated rationales.

\begin{figure*}[t]
    \centering
    \includegraphics[width=0.94\textwidth]{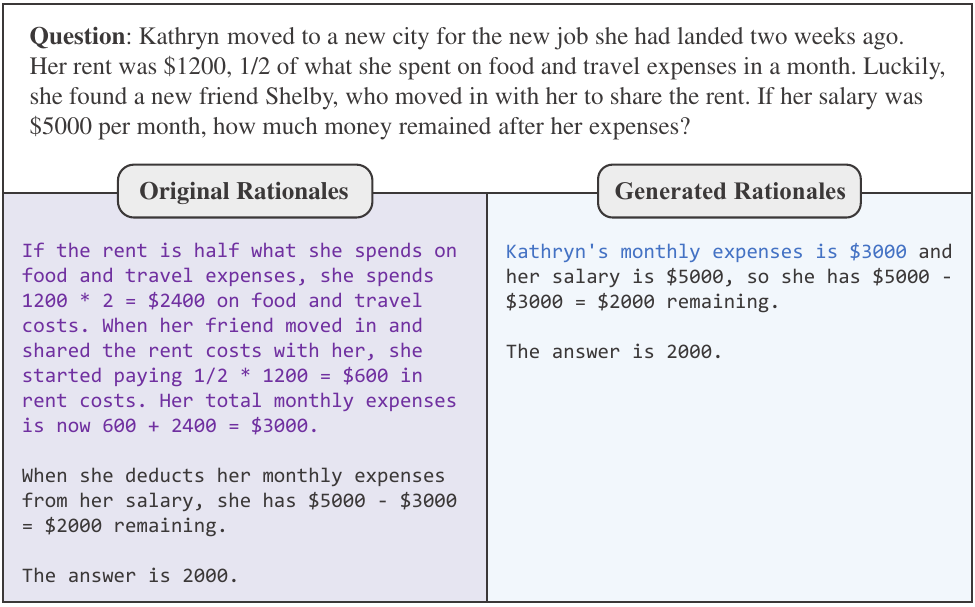}
    \caption{An example of \textbf{skipped steps} on the GSM8K dataset. The key steps within the original rationales are highlighted in \textcolor{Purple}{\textbf{purple}}, while the corresponding steps generated by Llama2 are indicated in \textcolor{Blue}{\textbf{blue}}. In this case, the model, guided by rationale-driven hints, skips critical steps and directly arrives at a premature conclusion.}
    \label{fig:error_1}
\end{figure*}

\begin{figure*}[t]
    \centering
    \includegraphics[width=0.94\textwidth]{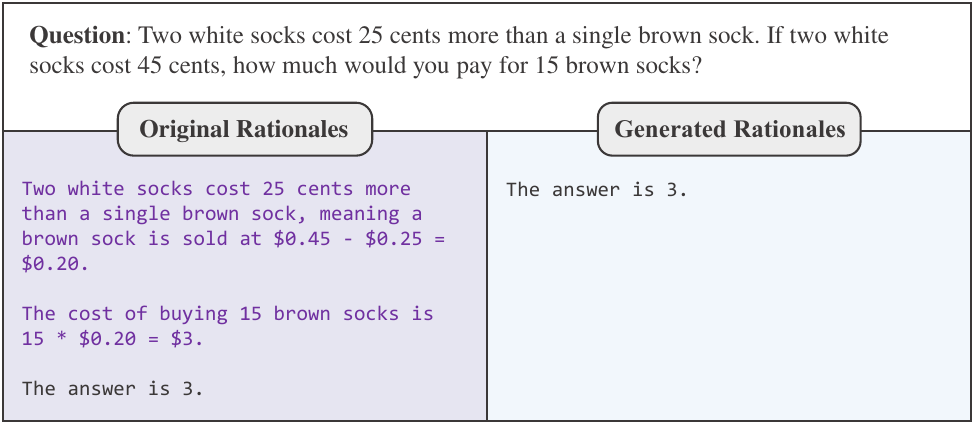}
    \caption{An example of \textbf{skipped steps} on the GSM8K dataset. The key steps within the original rationales are highlighted in \textcolor{Purple}{\textbf{purple}}. In this case, the model ignores all the steps and copies the final answer. Such cases arise mainly with answer-driven or rationale-driven hints.}
    \label{fig:error_2}
\end{figure*}

\paragraph{Hint Short-cutting.}
This occurs when the model is provided with the final answer or rationales as a hint during generation, causing it to skip steps \citep{star}. This leads to responses like those shown in Figure~\ref{fig:error_1} and Figure~\ref{fig:error_2}.

\begin{figure*}[t]
    \centering
    \includegraphics[width=0.94\textwidth]{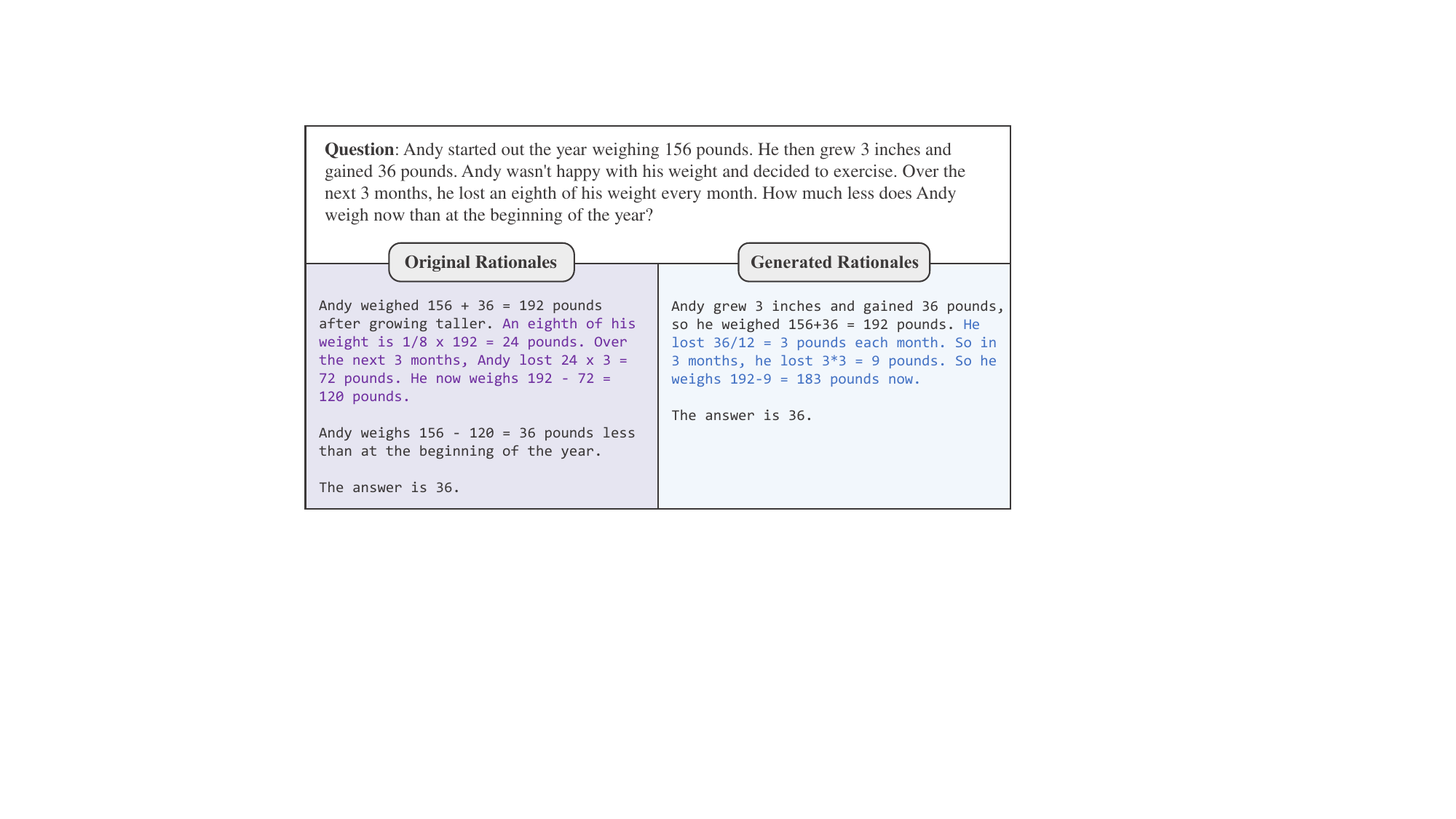}
    \caption{An example of a \textbf{hallucinated reasoning process} on the GSM8K dataset. The key steps within the original rationales are highlighted in \textcolor{Purple}{\textbf{purple}}, while the corresponding steps generated by Llama2 are indicated in \textcolor{Blue}{\textbf{blue}}. In this case, the model fails to generate the correct intermediate reasoning but still copies the correct final answer. This behavior can mislead the reward function and introduce noise into the training data in subsequent iterations. Such cases arise mainly with answer-driven hints.}
    \label{fig:error_3}
\end{figure*}

\paragraph{Spurious Correctness.}
This refers to cases where, when given the final answer as a hint, the model fails to generate the correct intermediate reasoning due to its limited capabilities but still copies the correct final answer and misleads the reward function. Responses classified as positive may contain incorrect or irrelevant rationales, and training on such data often incentivizes the model to overfit to spurious correlations, resulting in flat or even inverse scaling with more data during fine-tuning \citep{rl_on_incorrect}. 
Figure~\ref{fig:error_3} shows a case of spurious correct response.

\subsection{Case Study} \label{app:case_study}

\begin{figure*}[t]
    \centering
    \includegraphics[width=0.94\textwidth]{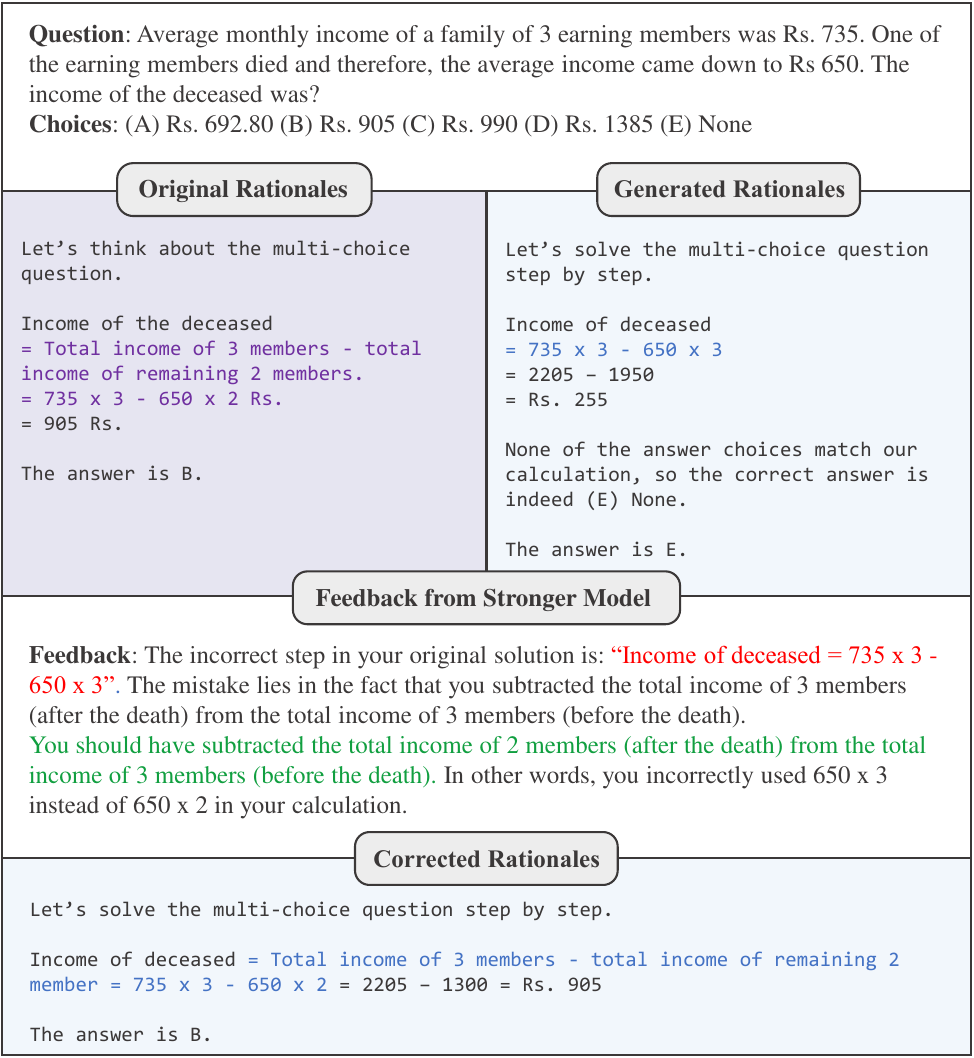}
    \caption{An example of \textbf{interactive sampling} on the AQuA dataset. The key steps within the original rationales are highlighted in \textcolor{Purple}{\textbf{purple}}, while the corresponding steps generated by Llama2 are indicated in \textcolor{Blue}{\textbf{blue}}. In the feedback from the stronger model, errors pointed out are marked in \textcolor{red}{\textbf{red}}, and corrections are indicated in \textcolor{Green}{\textbf{green}}. In this case, the model successfully corrected its mistake after receiving feedback and arrived at the correct final answer.}
    \label{fig:case_interact}
\end{figure*}

\begin{figure*}[t]
    \centering
    \includegraphics[width=0.94\textwidth]{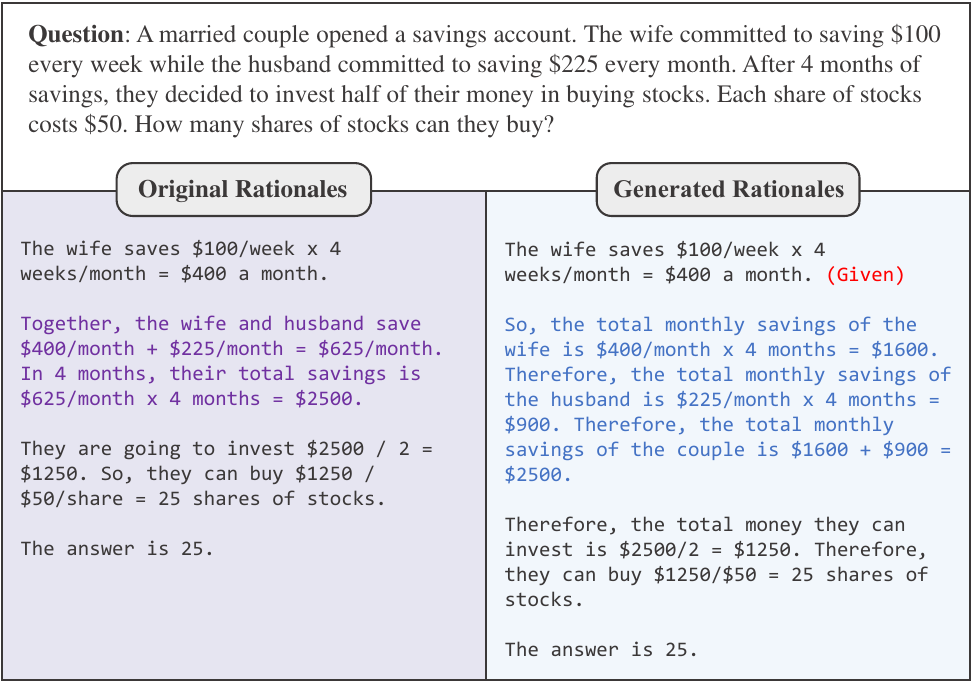}
    \caption{An example of \textbf{state reset} on the GSM8K dataset. The key steps within the original rationales are highlighted in \textcolor{Purple}{\textbf{purple}}, while the corresponding steps generated by Llama2 are indicated in \textcolor{Blue}{\textbf{blue}}. The \textcolor{red}{\texttt{(Given)}} highlights the partial rationale provided to the model as a hint. In this case, the model starts from the intermediate steps and gives a different solution, enriching the rationales in the training data.}
    \label{fig:case_state_reset}
\end{figure*}

Here, we present two successful cases: GSI (interactive sampling) in Figure~\ref{fig:case_interact} and GSI (state reset) in Figure~\ref{fig:case_state_reset}, demonstrating the effectiveness of our proposed method.

\subsection{Prompt Details} \label{app:prompt}

The prompt template of the vanilla self-improve method is presented in Figure \ref{prompt:vanilla}.
The prompt templates of proposed Guided Self-Improvement (\ours) are detailed in Figure \ref{prompt:answer}, \ref{prompt:rationale}, \ref{prompt:interact_teacher}, \ref{prompt:interact_student}, and \ref{prompt:reset}.

\begin{figure*}[t]
    \centering
\begin{tcolorbox}[colback=gray!5!white,colframe=gray!75!black]
\textbf{Input:}
\\~\\
Below is a math problem, please give a step-by-step answer. \\
\\
\#\#\# Question: \\
\texttt{\{question\}} \\
\\
\#\#\# Your step-by-step answer:
\end{tcolorbox}
    \caption{Prompt template for vanilla self-improve method. It is also used in the first step of \ours (Interactive Sampling), with the process transitioning into the subsequent interactive steps when incorrect results are sampled.}
\label{prompt:vanilla}
\end{figure*}

\begin{figure*}[t]
    \centering
\begin{tcolorbox}[colback=gray!5!white,colframe=gray!75!black]
\textbf{Input:}
\\~\\
Below is a math problem, please give a step-by-step answer. \\
\\
\#\#\# Question: \\
\texttt{\{question\}} \\
\texttt{\{answer\}} \\
\\
\#\#\# Your step-by-step answer:
\end{tcolorbox}
    \caption{Prompt template for \ours (Answer-driven) method.}
\label{prompt:answer}
\end{figure*}

\begin{figure*}[t]
    \centering
\begin{tcolorbox}[colback=gray!5!white,colframe=gray!75!black]
\textbf{Input:}
\\~\\
Below is a math problem with a reference answer. Using the reference answer as a guide, write your own answer. \\
\\
\#\#\# Question: \\
\texttt{\{question\}} \\
\\
\#\#\# Reference Answer: \\
\texttt{\{rationale\}} \\

\#\#\# Your detailed, complete and step-by-step answer:
\end{tcolorbox}
    \caption{Prompt template for \ours (Rationale-driven) method.}
\label{prompt:rationale}
\end{figure*}

\begin{figure*}[t]
    \centering
\begin{tcolorbox}[colback=gray!5!white,colframe=gray!75!black]
\textbf{Strong Model Input:}
\\~\\
You're a patient teacher who corrects mistakes and guides students, helping them find the correct answers on their own. For the following math problem, the original solution is incorrect. Please identify the incorrect step and explain why it is incorrect. \\
\\
\#\#\# Question: \\
\texttt{\{question\}} \\
\\
\#\#\# Student's original wrong answer: \\
\texttt{\{wrong\_answer\}} \\
\\
\#\#\# Correct reference answer: \\
\texttt{\{rationale\}} \\
\\
\#\#\# Your correction:
\end{tcolorbox}
    \caption{Prompt template for the strong model in \ours (Interactive Sampling) method.}
\label{prompt:interact_teacher}
\end{figure*}

\begin{figure*}[t]
    \centering
\begin{tcolorbox}[colback=gray!5!white,colframe=gray!75!black]
\textbf{Input:}
\\~\\
Below is a correction to your previous solution. Review this carefully and use it to revise your solution. Ensure that it includes all necessary steps clearly and thoroughly. \\
\\
\#\#\# Question: \\
\texttt{\{question\}} \\
\\
\#\#\# Your original wrong answer: \\
\texttt{\{wrong\_answer\}} \\
\\
\#\#\# Correction and guidance: \\
\texttt{\{correct\_message\}} \\
\\
\#\#\# Your revised, complete step-by-step solution:
\end{tcolorbox}
    \caption{Prompt template for the self-improved model in \ours (Interactive Sampling) method.}
\label{prompt:interact_student}
\end{figure*}

\begin{figure*}[t]
    \centering
\begin{tcolorbox}[colback=gray!5!white,colframe=gray!75!black]
\textbf{Input:}
\\~\\
Below is a math problem, please give a step-by-step answer. \\
\\
\#\#\# Question: \\
\texttt{\{question\}} \\
\\
\#\#\# Your step-by-step answer: \\
\texttt{\{partial\_rationale\}}
\end{tcolorbox}
    \caption{Prompt template for \ours (State Reset) method. The partial rationale is truncated from the complete one and placed at the end to guide the model in completing it.}
\label{prompt:reset}
\end{figure*}

\end{document}